\documentclass[11pt]{article}

\usepackage[final]{acl}

\usepackage{times}
\usepackage{latexsym}

\usepackage[T1]{fontenc}

\usepackage[utf8]{inputenc}

\usepackage{microtype}

\usepackage{inconsolata}

\usepackage{graphicx}

\usepackage{xspace}
\usepackage{enumitem}
\usepackage{amsmath}
\usepackage{bm}
\usepackage{multirow}
\usepackage{amssymb}
\usepackage{booktabs}
\usepackage{subcaption}
\usepackage{algorithm}
\usepackage{algorithmic} 
\usepackage{float}
\usepackage{xcolor}
\usepackage{colortbl}
\graphicspath{{figures/}}

\newcommand{\xb}{\mathbf{x}}
\newcommand{\yb}{\mathbf{y}}

\newcommand{\btheta}{\boldsymbol{\theta}}
\newcommand{\ours}{TPAW\xspace}
\usepackage{fontawesome5} 
\usepackage{seqsplit}
\title{Team-Based Self-Play With Dual Adaptive Weighting for \\ Fine-Tuning LLMs}

\author{
    Wu Li$^1$
    \quad Yigeng Zhou$^1$
    \quad Zesheng Shi$^1$
    \quad Yequan Wang$^2$
    \quad Min Zhang$^1$
    \quad  \textbf{Jing Li}$^1$\textsuperscript{\texorpdfstring{\faIcon[regular]{envelope}}{}} 
    \\$^{1}$Harbin Institute of Technology, Shenzhen, China  \\
    $^{2}$Beijing Academy of Artificial Intelligence, Beijing, China
   \\
    \texttt{li-555@outlook.com} \quad \texttt{jingli.phd@hotmail.com}  
}

\begin{document}
\maketitle
\begin{abstract}
While recent self-training approaches have reduced reliance on human-labeled data for aligning LLMs, they still face critical limitations: (i) sensitivity to synthetic data quality, leading to instability and bias amplification in iterative training; (ii) ineffective optimization due to a diminishing gap between positive and negative responses over successive training iterations. In this paper, we propose Team-based self-Play with dual Adaptive Weighting (TPAW), a novel self-play algorithm designed to improve alignment in a fully self-supervised setting. TPAW adopts a team-based framework in which the current policy model both collaborates with and competes against historical checkpoints, promoting more stable and efficient optimization. To further enhance learning, we design two adaptive weighting mechanisms: (i) a response reweighting scheme that adjusts the importance of target responses, and (ii) a player weighting strategy that dynamically modulates each team member’s contribution during training. Initialized from a SFT model, TPAW iteratively refines alignment without requiring additional human supervision. Experimental results demonstrate that TPAW consistently outperforms existing baselines across various base models and LLM benchmarks. Our code is publicly available at \url{https://github.com/lab-klc/TPAW} 

\let\thefootnote\relax\footnotetext{\faIcon[regular]{envelope}~Corresponding author.}	
\end{abstract}

\section{Introduction}

Large Language Models (LLMs) have demonstrated remarkable capabilities across a wide range of NLP tasks~\citep{radford2019language,achiam2023gpt}, driven by pre-training on massive datasets and further refined through alignment techniques such as Supervised Fine-Tuning (SFT), Reinforcement Learning from Human Feedback (RLHF)~\citep{ouyang2022training,christiano2017deep,stiennon2020learning}, and Direct Preference Optimization (DPO)~\citep{rafailov2023direct}.
Despite their success, these alignment methods heavily depend on human-annotated data, which limits their scalability. To address this, recent research has explored self-training paradigms~\citep{DBLP:conf/acl/WangKMLSKH23,DBLP:conf/icml/YuanPCLSXW24,chen2024self} that utilize model-generated data. 
Among them, methods such as Self-Play Fine-Tuning (SPIN)~\citep{chen2024self} have shown promise by repurposing existing SFT datasets to iteratively align models without the need for additional human annotations.

However, these self-training approaches still face several limitations that may hinder their alignment performance. 
First, based on current model's performance, self-rewarding or self-play mechanisms fail to fully utilize the model’s accumulated training trajectory throughout the iterative training process. In addition, these approaches exhibit training instability and high sensitivity to the quality of synthetic data, often leading to the accumulation and amplification of errors and biases during self-training~\citep{ren2024bias,xu2024pride}.
Second, most self-training methods use DPO-like objectives that jointly optimize positive and negative samples. However, studies have shown that the probabilities of both $y^+$ and $y^-$ often decrease during training~\citep{pang2024iterative,yuan2025advancing,rafailov2024from,tajwar2024preference}, possibly due to the high similarity between them~\citep{pal2024smaug, razin2025unintentional}. This problem is exacerbated in self-training methods, where the distinction between positive and negative responses diminishes over iterations~\citep{song2025mind, huang2024self}, resulting in noisy preference data and suboptimal policy updates~\citep{wang2025cream}.

\begin{figure*}[t]
  \centering
  \includegraphics[width=0.97\textwidth]{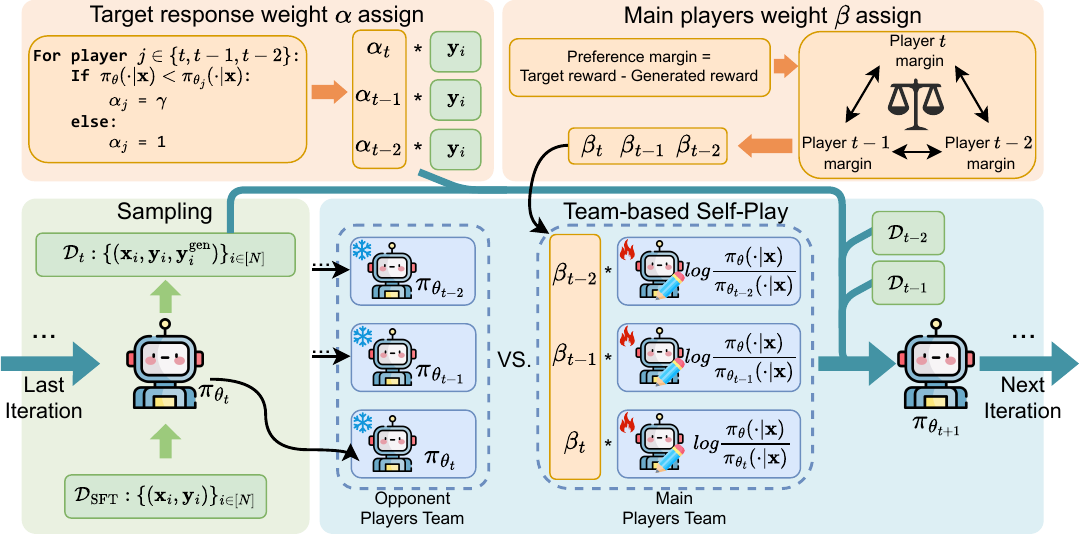}
  \caption{The workflow of \ours. During response sampling, the model $\pi_{\btheta_t}$ generates responses $\yb^{\text{gen}}$ from SFT prompts $\xb$, forming a triple dataset $\mathcal{D}_t$. Next, \ours adaptively assigns target response weights $\bm{\alpha}$ based on the target response likelihood, and main player weights $\bm{\beta}$ based on the reward margin between the target and generated responses. Finally, the policy model $\pi_{\btheta}$ is fine-tuned using the overall Team-based Self-Play training objective, resulting in model $\pi_{\btheta_{t+1}}$ for next iteration.}
  \vspace{-10pt}
  \label{fig:method}
\end{figure*}

To address these challenges, we propose \textbf{T}eam-based self-\textbf{P}lay with dual \textbf{A}daptive \textbf{W}eighting (\ours), as illustrated in Figure~\ref{fig:method}.
\ours consists of two key components: (1) a team-based self-play framework that optimizes policy model by leveraging historical checkpoints, leading to more stable and efficient training; and (2) two adaptive weighting mechanisms—one for assessing the importance of target responses and the other for adjusting the influence of each player in team-based interactions—both of which contribute to improved model performance and better alignment with the target response distribution.
Starting from a SFT model, \ours iteratively trains the LLM using SFT dataset. In each iteration, historical policy checkpoints act as opponents, generating responses aligned with the target distribution. 
Concurrently, the current policy model, in collaboration with these historical checkpoints, forms the reward model, which acts as the main player team that works to distinguish authentic target responses from LLM-generated content.
During this process, the two adaptive weighting mechanisms dynamically adjust (i) the contribution of each main player and (ii) the weight of target responses, ensuring effective learning and adaptation.

In summary, the key contributions of our paper are as follows:

\vspace{-3mm}
\begin{itemize}[left=0cm]
    \setlength{\itemsep}{3pt}  
    \setlength{\parskip}{0pt}  
    \setlength{\parsep}{0pt}   
    \item \textbf{Team-Based Self-Play Framework}: We propose the first self-play method that unifies historical policy checkpoints into both the opponent and main player teams, overcoming the limitations of multi-iteration self-training methods by fully optimizing on historical training trajectories.
    \item \textbf{Dual Adaptive Weighting Mechanism}: We propose a dual adaptive weighting scheme that jointly optimizes two key objectives: (i) balancing the main players’ discriminative ability with LLM's alignment to the target response distribution through adaptive response reweighting, and (ii) adaptively assigning influence weights to the main players based on their discriminative strengths.
    \item \textbf{Theoretical and Empirical Advancements}: Empirically, \ours significantly enhances the well-aligned SFT model, outperforming baselines trained on similar datasets across multiple benchmarks. Further analysis confirms that \ours better exploits the SFT dataset and achieves closer alignment with the target response distribution.
\end{itemize}

\section{Related Work}

\paragraph{Self-play.}
Self-play~\citep{samuel1959some,tesauro1995temporal} is a training methodology in which a player repeatedly interacts with copies of itself to improve performance through exploration, evaluation, and refinement of strategies. This iterative process enables the discovery of increasingly effective solutions within a given environment.
A landmark example of self-play research is AlphaGo Zero~\citep{silver2017mastering}, which, unlike previous models, does not rely on human data but instead leverages self-play combined with Monte Carlo Tree Search, ultimately achieving performance that exceeds human capabilities.
Recent studies have demonstrated the application of self-play in LLMs~\citep{wu2024self,zhao2025absolute,cheng2024self,swamy2024minimaximalist}. SPIN~\citep{chen2024self} enables LLMs to iteratively generate training data and refine their strategies by distinguishing between self-generated and human-annotated responses. This approach yields substantial performance gains without requiring additional human-labeled data.
AutoIF~\citep{dong2024self} improves the performance of LLMs on complex tasks by automatically generating instruction-following data and utilizing code verification and execution feedback-based rejection sampling.
However, these methods fail to fully utilize historical checkpoints and are highly sensitive to the quality of synthetic data, which makes it difficult for their outputs to fit the target response distribution and amplifies errors and biases during the iteration process.

\vspace{-3mm}
\paragraph{Alignment.}
Reinforcement Learning from Human Feedback (RLHF)~\citep{ouyang2022training, DBLP:conf/icml/0001PMMFLBHCRP24} aims to align LLMs with human preferences. This method first learns a reward model that reflects human preferences and then uses reinforcement learning algorithms to maximize the reward.
The RLHF method involves reward modeling, distributed training of multiple models, and expert annotations, presenting operational challenges. Therefore, recent research has focused on reducing or eliminating reliance on reward models to simplify the RLHF process~\citep{rafailov2023direct,DBLP:journals/corr/abs-2402-01306, DBLP:conf/nips/0001X024}.
Alignment research has also extended to safety settings~\citep{shi2025safety}.
DPO (Direct Policy Optimization)~\citep{rafailov2023direct} is an improved method that directly optimizes the policy instead of relying on traditional reward models, effectively reducing the dependence on complex reward modeling and enhancing training efficiency and stability.
Building upon similar ideas, \ours directly trains the model to distinguish ``good'' responses from the SFT dataset and ``bad'' responses generated by the model, significantly reducing the reliance on human-annotated data.

\section{Methodology}

In this section, we present \ours, a novel approach that enhances LLM alignment through team-based self-play with dual adaptive weighting, as shown in Figure~\ref{fig:method}. 
First, we formulate the Team-Based Self-Play framework, establishing its theoretical foundations and operational structure. 
Next, we introduce the Dual Adaptive Weighting mechanism, which dynamically adjusts both the importance of training target responses and the influence of main players during the team-based self-play process. 
Finally, we provide a detailed algorithm for the practical implementation of \ours, including pseudocode and optimization strategies.
\subsection{Team-Based Self-Play}
Conventional self-training methods typically rely on a single model’s performance, suffering from training instability and are highly sensitive to the quality of synthetic data, which can lead to the propagation and amplification of errors and biases over successive iterations~\citep{ren2024bias,xu2024pride}.
To overcome these limitations, we revisit the self-play paradigm by effectively leveraging the historical checkpoints accumulated across iterative training.

Based on this idea, we propose a Team-Based Self-Play framework that reframes self-play as a competitive game between two teams. In this setting, the main players are responsible for distinguishing between human-written and LLM-generated responses, while the opponents aim to generate responses that are as human-like as possible. Both teams are composed of instances of the same LLM, but at different training stages: the opponents are drawn from earlier checkpoints, whereas the main players are composites built from the current model and its previous iterations.

This team-based dynamic not only introduces diversity and adversarial robustness into the training process but also allows for targeted optimization—since some main players may struggle to distinguish responses effectively, we can adaptively refine their learning to improve overall training efficiency. Furthermore, the inclusion of the training trajectory serves as an implicit form of regularization, helping to stabilize the learning process and mitigate error accumulation over time.

Our Team-Based Self-Play framework proceeds in each iteration with two key steps: $(1)$ sampling from opponents, and $(2)$ updating the main players.

\paragraph{Sampling from Opponents.}
At iteration $t+1$, the opponent team is composed of the LLMs from the three most recent iterations, denoted as $(\pi_{\btheta_{t}},\pi_{\btheta_{t-1}},\pi_{\btheta_{t-2}})$. This ensures a dynamic balance between recent and historical strategies during training. 
For the newly added opponent $\pi_{\btheta_{t}}$ in this iteration, we use prompts $\xb$ from the SFT dataset $\mathcal{D}_{\text{SFT}}$ to generate responses $\yb^{\text{gen}}$ according to the $\pi_{\btheta_t}(\cdot|\xb)$.
Then we pair these generated responses with the original dataset entries $(\mathbf{x}_i,\mathbf{y}_i) \in \mathcal{D}_{\text{SFT}}$ to construct a new triple-form dataset: $\mathcal{D}_t = \{(\mathbf{x}_i, \mathbf{y}_i, \mathbf{y}_i^{\text{gen}})\}_{i=1}^{N}$.
In addition, to construct the complete sampling dataset for the opponent team, denoted as $\mathcal{D}_O$, we aggregate the triple datasets from the most recent three iterations:
$\mathcal{D}_O = \mathcal{D}_t \cup \mathcal{D}_{t-1} \cup \mathcal{D}_{t-2}$.

\paragraph{Forming Main Players Team.} 
The main players are designed to be trained to distinguish between responses generated by LLMs and those written by humans.  
Inspired by SPIN~\citep{chen2024self} and DPO~\citep{rafailov2023direct}'s implicit reward:
\begin{align}
\label{eq:implicitreward}
r(\xb, \yb) = \lambda\cdot \log \frac{\pi_{r}(\yb | \xb)}{\pi_{\text{ref}}(\yb | \xb)}+\lambda\cdot \log Z(x),
\end{align}

we define each main player as:

\begin{align}
\label{eq:mainplayer}
P_j(\xb, \yb) = \lambda\cdot \log \frac{\pi_{\btheta}(\yb | \xb)}{\pi_{\mathrm{\btheta_{j}}}(\yb | \xb)},~j \in \{t, t-1, t-2\} ,
\end{align}

where $\lambda > 0$ is a regularization parameter used to constrain the deviation of $\pi_{\btheta_{t+1}}$ from $\pi_{\btheta_{t}}$, thereby stabilizing the self-play process.

In this context, as we are only concerned with different responses corresponding to the same prompt $\xb$, the normalization term $\lambda\cdot \log Z(x)$ in Eq.~\ref{eq:implicitreward} can be omitted. 
Consequently, Eq.~\ref{eq:mainplayer} can be interpreted as an implicit reward score, jointly inferred by the current model and a historical model, conditioned on the given prompt $\xb$ and response $\yb$.
Within our framework, this score—produced by the main player—quantifies how well the response $\yb$ aligns with the target distribution.
Ideally, main players should assign higher scores to responses from the supervised fine-tuning dataset ($\yb \in \mathcal{D}_{\text{SFT}}$) and lower scores to model-generated responses ($\yb^{\text{gen}} \sim \pi_{\btheta_j}(\cdot \mid \xb)$).

Furthermore, to enable each player $P_j$ to learn the distinction between $\yb$ and $\yb^{\text{gen}}$, the single main player's training objective becomes:
\begin{align}
\label{eq:optim}
\mathcal{L}_{j} = -\mathbb{E}_{\mathcal{D}_O}\big[\ell\big(P_j(\xb, \yb)-P_j(\xb, \yb^{\text{gen}})\big)\big].
\end{align}

Following SPIN~\citep{chen2024self}, we use the logistic loss $\ell(t) := \log(1 + \exp(-t))$ due to its non-negativity, smoothness, and exponential tail decay as $t \to \infty$. This choice helps prevent excessive growth in the magnitude of $P_j$. 

In subsequent subsections, we will detail how to further optimize each player’s objective based on Eq.~\ref{eq:optim}, and describe how to aggregate these individual objectives into a unified team objective using a dual adaptive weighting mechanism.

\subsection{Dual Adaptive Weighting}

To further enhance the robustness and alignment of the team-based self-play framework, we introduce a dual adaptive weighting mechanism that dynamically adjusts the importance of training target responses and the influence of main players during team-based self-play.

\paragraph{Adaptive Target Response Weighting.}
Most existing self-training methods adopt optimization objectives similar to DPO, simultaneously optimizing both positive and negative samples.
However, recent studies have shown that when the similarity between positive and negative responses is high~\citep{pal2024smaug,razin2025unintentional}, or the probability of generating negative responses is low~\citep{ren2025learning}, the model's likelihood of producing preferred (i.e., target) responses can deteriorate during training.
We also find the decrease phenomenon in our ablation experiment and SPIN baseline in Section~\ref{sec:analyse}.

To address this issue, we propose Adaptive Target Response Weighting, a method designed to balance the discriminative capacity of the main players with the alignment of the LLM policy to the target distribution.
Specifically, based on each player $P_j$’s optimization objective in Eq.~\ref{eq:optim}, our method adaptively increases the reward score $P_j(\xb, \yb)$ for the target response $\yb$ when the model's likelihood for $\yb$ decreases relative to a reference model. This adjustment effectively increases the weight of the target response during training, encouraging the model to better refit to the target distribution.

We define the adaptive target response weight for player $P_j$ as:
\begin{align}
\label{eq:target_weight}
\bm{\alpha}_{j} = \left\{
\begin{array}{ll}
\bm{\eta} & \text{if } P_j(\xb, \yb) \leq 0,(\mathbf{x}, \mathbf{y}) \in \mathcal{D}_O,  \\
1   & \text{else},
\end{array}
\right.
\end{align}
where $\bm{\eta} > 1$ is a hyperparameter that controls the degree of weight growth when the target response score is low.

Incorporating this adaptive weighting into the player's objective, we modify Eq.~\ref{eq:optim} as follows:

\begin{align}
\label{eq:optim2}
\mathcal{L}_{j}=-\mathbb{E}_{\mathcal{D}_O}\big[\ell\big(\bm{\alpha}_{j} \cdot P_j(\xb, \yb)-P_j(\xb, \yb^{\text{gen}})\big)\big],
\end{align}

\paragraph{Adaptive Main Players Weighting.} 
In the Team-Based Self-Play framework, multiple main players are introduced to fully leverage the potential of historical model checkpoints in enhancing the discriminative capabilities of the main players.
However, in iteration $t+1$, since $P_{t-1}$ and $P_{t-2}$ have already undergone optimization in previous iterations, statically assigning weights to the main players can lead to suboptimal training. 
Specifically, $P_t$ may be undertrained, while $P_{t-1}$ and $P_{t-2}$ risk overfitting.

To address this, we propose Adaptive Main Player Weighting, which dynamically assigns greater weight to the main player when it exhibits weaker discriminative performance in distinguishing between human-written and LLM-generated responses.
Specifically, for a given player $P_{j}$, we define the preference margin $m_j$ as:
\begin{align}
\label{eq:margin}
m_j = P_j(\xb, \yb)-P_j(\xb, \yb^{\text{gen}}).
\end{align}

As previously discussed, $P_j$ provides a reward score indicating how well the response $\yb$ aligns with the target distribution. 
Thus, the preference margin $m_j$ quantifies the main player's effectiveness in discriminating between human-written and LLM-generated responses for a given sample.

We then apply the softmax function to compute each player's weight $\bm{\beta}_{j}$ based on their margin value:
\begin{align}
\label{eq:softmax}
\bm{\beta}_{j} = \frac{e^{-\bm{\gamma} \cdot m_j}}{\displaystyle\sum_{k\in \text{main players}}e^{-\bm{\gamma} \cdot m_k}}.
\end{align}

For each main player, a smaller margin—indicating poorer judgment on a sample—results in a higher weight for that sample. The hyperparameter $\bm{\gamma}$ controls the sensitivity of the weight distribution to the margin values. A larger $\bm{\gamma}$ amplifies the influence of small margins, allowing them to dominate the softmax output. Conversely, as $\bm{\gamma}$ approaches zero, the weights converge to a uniform distribution, effectively ignoring differences in margin.
To suppress the influence of extremely small values and unstable noise, we set weights smaller than 0.01 to zero and re-normalize the remaining weights.

\paragraph{Overall Training Objective.}
By assigning weights to each main player, we can aggregate their individual objectives from Eq.~\ref{eq:optim2} into a unified overall training objective for the entire main player team:

\begin{equation}
\mathcal{L}_{\text{ours}} = \sum_{i=t-2}^{t} \bm{\beta}_{i} \cdot \mathcal{L}_{i}(\pi;\pi_i)
\end{equation}

\begin{algorithm}[t]
\renewcommand{\algorithmicrequire}{\textbf{Input:}}
\renewcommand{\algorithmicensure}{\textbf{Output:}}
\footnotesize
\algsetup{indent=6pt}
\caption{Team-based self-Play with dual Adaptive Weighting (\ours)}
\label{alg:main}
\vspace{-0.2em}
\begin{algorithmic}[0]

\REQUIRE SFT dataset $\mathcal{D}_{\text{SFT}} = \{(\mathbf{x}_i, \mathbf{y}_i)\}_{i=1}^{N}$; Initial SFT policy $\pi_{\bm{\theta}_0}$; Number of iterations $T$.

\FOR{$t = 0$ to $T-1$}
    \STATE Generate responses: $\{\mathbf{y}_i^{\text{gen}}\}_{i=1}^{N} \sim \pi_{\bm{\theta}_t}(\cdot \mid \mathbf{x}_i),~ \forall \mathbf{x}_i \in \mathcal{D}_{\text{SFT}}$
    
    \STATE Construct pairwise preference dataset: 
    
    ~~~~~~~~~~~~~~$\mathcal{D}_t = \{(\mathbf{x}_i, \mathbf{y}_i, \mathbf{y}_i^{\text{gen}})\}_{i=1}^{N}, ~(\mathbf{x}_i,\mathbf{y}_i) \in \mathcal{D}_{\text{SFT}}$

    \STATE Aggregate past opponents: 
    $\mathcal{D}_O = \mathcal{D}_t \cup \mathcal{D}_{t-1} \cup \mathcal{D}_{t-2}$
    
    \FOR{$j \in \{t, t-1, t-2\}$}
        \STATE Get target response weights $\bm{\alpha}_j$ according to Eq.~\ref{eq:target_weight}
        \STATE Compute each player's loss for each sample:

{
\vspace{-1.8em}
\tiny
\begin{equation*}
\bm{l}_{ij}(\pi_{\bm{\theta}};\pi_{\bm{\theta}_j}) =\ell \left( \bm{\alpha}_j \log \frac{\pi_{\bm{\theta}}(\mathbf{y}_i \mid \mathbf{x}_i)}{\pi_{\bm{\theta}_j}(\mathbf{y}_i \mid \mathbf{x}_i)} - \log \frac{\pi_{\bm{\theta}}(\mathbf{y}_i^{\text{gen}} \mid \mathbf{x}_i)}{\pi_{\bm{\theta}_j}(\mathbf{y}_i^{\text{gen}} \mid \mathbf{x}_i)} \right),
\end{equation*}
\vspace{-2em}
}
        
        \STATE where $(\mathbf{x}_i, \mathbf{y}_i, \mathbf{y}_i^{\text{gen}}) \in \mathcal{D}_O.$
    \ENDFOR

    \STATE Determine policy weighting coefficients: $\bm{\beta}_t, \bm{\beta}_{t-1}, \bm{\beta}_{t-2}$ according to Eq.~\ref{eq:softmax}

    \STATE Update policy parameters:
\vspace{-1.2em}
{\scriptsize
\begin{equation*}
\bm{\theta}_{t+1} = \arg\min_{\bm{\theta} \in \bm{\Theta}} 
\sum_{i=1}^{3N} \lambda \cdot 
\left(
    \bm{\beta}_t \bm{l}_{i,t} + 
    \bm{\beta}_{t-1} \bm{l}_{i, t-1} + 
    \bm{\beta}_{t-2} \bm{l}_{i, t-2}
\right)
\end{equation*}
}
\vspace{-1.5em}
\ENDFOR

\ENSURE Final aligned policy $\pi_{\bm{\theta}_T}$.
\end{algorithmic}
\vspace{-0.3em}
\end{algorithm}

\subsection{Summary of TPAW}

With the proposed Team-Based Self-Play framework and Dual Adaptive Weighting mechanism in place, we now present the implementation of \ours in Algorithm~\ref{alg:main}. An overview of the entire framework is shown in Figure~\ref{fig:method}.

At each iteration, the current policy generates responses on the SFT dataset and forms preference pairs with the target responses. These are combined with data from previous iterations to construct an opponent dataset. 
For each past policy, we compute adaptive response weights $\bm{\alpha}_j$ to guide learning at the sample level, and player weights $\bm{\beta}_j$ to balance the influence of different main players also at the sample level.
The policy is then updated by minimizing a weighted sum of losses against each opponent, enabling alignment through iterative, adaptive team-based self-play.

Specifically, at $t=0$, we ignore the historical models $\pi_{t-1}$ and $\pi_{t-2}$, at $t=1$, we ignore $\pi_{t-2}$. To better illustrate how policies evolve over time, Appendix~\ref{app:sequence} summarizes the training process for each policy, including initialization, training data, opponents, and main players.

\begin{table*}[t]
    \huge
    \centering
    \setlength{\tabcolsep}{6pt} 
    \renewcommand{\arraystretch}{1.5}
    \resizebox{\textwidth}{!}{%
    \begin{tabular}{c | c| c c c c c c | c || c c c c c c | c}
    \toprule
        \rowcolor[gray]{0.9}
        & & \multicolumn{6}{c|}{\textbf{Open LLM Leaderboard V1}} & & \multicolumn{6}{c|}{\textbf{Open LLM Leaderboard V2}} &\\[0.5ex]
        \rowcolor[gray]{0.9}
        \multirow{-2}{*}{\textbf{Model}} & \multirow{-2}{*}{\textbf{Method}}
        & Arc & TQA & Wino & GSM8k & HS & MMLU & \multirow{-2}{*}{ \phantom{0} \textbf{V1 Avg.} \phantom{0}} 
        & IFEval & BBH & Math & GPQA & MUSR & MMLU-p & \multirow{-2}{*}{ \phantom{0} \textbf{V2 Avg.} \phantom{0}} \\
    \midrule
         & SFT & 50.94 & 46.83 & 64.80 & 51.25 & 64.91 & 58.93 & 56.28 
                                 & 31.73 & 15.15 & 6.65 & 3.47 & \textbf{6.02} & 17.38 & 13.40 \\
    \cline{2-16}
            & DPO & 54.52 & 49.39 & 64.96 & 56.71 & 66.29 & 59.41 & 58.55 
                                 & 31.79 & 16.25 & 9.89 & 2.24 & 2.57 & 18.24 & 13.50 \\
    \cline{2-16}
          &\texttt{SPIN} Iter-1 & 51.79 & 47.82 & 64.33 & \underline{53.83} & 65.76 & 59.26 & 57.13 
                             & 32.44 & 15.47 & 8.23 & 4.03 & 5.55 & 17.57 & 13.88 \\
          &\texttt{SPIN} Iter-2 & 52.05 & 48.25 & 64.40 & 53.45 & 66.00 & 59.43 & 57.26 
                             & 33.97 & 15.22 & 8.99 & 3.69 & 5.65 & 17.52 & 14.17 \\
          &\texttt{SPIN} Iter-3 & 52.65 & \underline{48.38} & 64.48 & 53.53 & \underline{66.25} & 59.56 & 57.47 
                             & 35.34 & 14.44 & 9.06 & \textbf{4.47} & 5.16 & 17.58 & 14.34 \\
          &\texttt{SPIN} Iter-4 & \underline{53.58} & \textbf{48.41} & 64.72 & 53.30 & 66.05 & \underline{59.58} & 57.61 
                             & 33.82 & 14.33 & 9.21 & 4.25 & 5.70 & 17.54 & 14.14 \\
    \cline{2-16}
          &\cellcolor[HTML]{FEF7F4}\ours Iter-1~(Ours) & \cellcolor[HTML]{FEF7F4}52.99 & \cellcolor[HTML]{FEF7F4}47.64 & \cellcolor[HTML]{FEF7F4}64.96 & \cellcolor[HTML]{FEF7F4}52.77 & \cellcolor[HTML]{FEF7F4}65.77 & \cellcolor[HTML]{FEF7F4}59.50 & \cellcolor[HTML]{FEF7F4}57.27 
                             & \cellcolor[HTML]{FEF7F4}32.35 & \cellcolor[HTML]{FEF7F4}\textbf{15.60} & \cellcolor[HTML]{FEF7F4}8.69 & \cellcolor[HTML]{FEF7F4}3.24 & \cellcolor[HTML]{FEF7F4}5.66 & \cellcolor[HTML]{FEF7F4}17.65 & \cellcolor[HTML]{FEF7F4}13.87 \\
          &\cellcolor[HTML]{FCF0EA}\ours Iter-2~(Ours) & \cellcolor[HTML]{FCF0EA}53.41 & \cellcolor[HTML]{FCF0EA}47.91 & \cellcolor[HTML]{FCF0EA}64.17 & \cellcolor[HTML]{FCF0EA}53.15 & \cellcolor[HTML]{FCF0EA}66.08 & \cellcolor[HTML]{FCF0EA}\underline{59.58} & \cellcolor[HTML]{FCF0EA}57.38 
                             & \cellcolor[HTML]{FCF0EA}35.52 & \cellcolor[HTML]{FCF0EA}15.37 & \cellcolor[HTML]{FCF0EA}8.69 & \cellcolor[HTML]{FCF0EA}\underline{4.36} & \cellcolor[HTML]{FCF0EA}\underline{5.98} & \cellcolor[HTML]{FCF0EA}17.72 & \cellcolor[HTML]{FCF0EA}14.61 \\
          &\cellcolor[HTML]{FBEAE1}\ours Iter-3~(Ours) & \cellcolor[HTML]{FBEAE1}\textbf{53.67} & \cellcolor[HTML]{FBEAE1}47.91 & \cellcolor[HTML]{FBEAE1}\underline{65.04} & \cellcolor[HTML]{FBEAE1}53.60 & \cellcolor[HTML]{FBEAE1}66.18 & \cellcolor[HTML]{FBEAE1}59.52 & \cellcolor[HTML]{FBEAE1}\underline{57.65} 
                             & \cellcolor[HTML]{FBEAE1}\textbf{36.10} & \cellcolor[HTML]{FBEAE1}15.03 & \cellcolor[HTML]{FBEAE1}\textbf{10.20} & \cellcolor[HTML]{FBEAE1}4.14 & \cellcolor[HTML]{FBEAE1}5.43 & \cellcolor[HTML]{FBEAE1}\textbf{18.03} & \cellcolor[HTML]{FBEAE1}\textbf{14.82} \\
          \multirow{-9}{*}{\textbf{Qwen2.5-1.5B}}  &\cellcolor[HTML]{F9E4D9}\ours Iter-4~(Ours) & \cellcolor[HTML]{F9E4D9}52.56 & \cellcolor[HTML]{F9E4D9}47.90 & \cellcolor[HTML]{F9E4D9}\textbf{65.11} & \cellcolor[HTML]{F9E4D9}\textbf{55.04} & \cellcolor[HTML]{F9E4D9}\textbf{66.30} & \cellcolor[HTML]{F9E4D9}\textbf{59.64} & \cellcolor[HTML]{F9E4D9}\textbf{57.76} 
                             & \cellcolor[HTML]{F9E4D9}\underline{36.04} & \cellcolor[HTML]{F9E4D9}\underline{15.51} & \cellcolor[HTML]{F9E4D9}\underline{9.37} & \cellcolor[HTML]{F9E4D9}\underline{4.36} & \cellcolor[HTML]{F9E4D9}4.73 & \cellcolor[HTML]{F9E4D9}\underline{17.82} & \cellcolor[HTML]{F9E4D9}\underline{14.64} \\
    \bottomrule
    \toprule
         & SFT & 60.75 & 51.65 & 74.82 & 53.15 & 79.48 & 63.72 & 63.93   & 36.34 & 26.28 & 4.61 & 6.15 & 11.77 & 21.01 & 17.69 \\
    \cline{2-16}
         & DPO & 63.14 & 57.42 & 74.35 & 50.27 & 80.46 & 63.08 & 64.79   & 36.83 & 25.83 & 4.53 & 6.60 & 11.78 & 21.91 & 17.91 \\
    \cline{2-16}
          &\texttt{SPIN} Iter-1 & 61.69 & 53.62 & 75.3 & 53.9 & 80.44 & 63.94 & 64.82   & 42.91 & 28.05 & 4.83 & \underline{6.26} & 13.17 & 22.84 & 19.68 \\
          &\texttt{SPIN} Iter-2 & 62.03 & 53.81 & 75.61 & 52.31 & 80.63 & 63.97 & 64.73   & 43.21 & 28.17 & 5.14 & 4.81 & 13.28 & 22.93 & 19.59 \\
          &\texttt{SPIN} Iter-3 & 62.2 & 53.98 & 76.01 & 51.71 & 80.74 & 64.15 & 64.80   & 44.13 & \textbf{28.46} & 5.21 & 3.36 & 12.96 & 22.92 & 19.51 \\
          &\texttt{SPIN} Iter-4 &  62.03 & 53.92 & 76.01 & 52.54 & \underline{80.78} & 63.97 & 64.88   & 44.01 & \underline{28.36} & 4.98 & 3.8 & 13.44 & 23.02 & 19.60 \\
    \cline{2-16}
          &\cellcolor[HTML]{FEF7F4}\ours Iter-1~(Ours) & \cellcolor[HTML]{FEF7F4}62.88 & \cellcolor[HTML]{FEF7F4}53.71 & \cellcolor[HTML]{FEF7F4}75.93 & \cellcolor[HTML]{FEF7F4}55.19 & \cellcolor[HTML]{FEF7F4}80.66 & \cellcolor[HTML]{FEF7F4}64.29 & \cellcolor[HTML]{FEF7F4}65.44 
                             & \cellcolor[HTML]{FEF7F4}38.76 & \cellcolor[HTML]{FEF7F4}28.16 & \cellcolor[HTML]{FEF7F4}5.21 & \cellcolor[HTML]{FEF7F4}6.15 & \cellcolor[HTML]{FEF7F4}13.65 & \cellcolor[HTML]{FEF7F4}23.17 & \cellcolor[HTML]{FEF7F4}19.18 \\
          &\cellcolor[HTML]{FCF0EA}\ours Iter-2~(Ours) & \cellcolor[HTML]{FCF0EA}62.97 & \cellcolor[HTML]{FCF0EA}54.11 & \cellcolor[HTML]{FCF0EA}76.16 & \cellcolor[HTML]{FCF0EA}\textbf{55.50} & \cellcolor[HTML]{FCF0EA}80.63 & \cellcolor[HTML]{FCF0EA}64.45 & \cellcolor[HTML]{FCF0EA}65.64 
                             & \cellcolor[HTML]{FCF0EA}42.43 & \cellcolor[HTML]{FCF0EA}27.85 & \cellcolor[HTML]{FCF0EA}\underline{6.19} & \cellcolor[HTML]{FCF0EA}\textbf{6.38} & \cellcolor[HTML]{FCF0EA}\textbf{16.22} & \cellcolor[HTML]{FCF0EA}23.02 & \cellcolor[HTML]{FCF0EA}20.35 \\
          &\cellcolor[HTML]{FBEAE1}\ours Iter-3~(Ours) & \cellcolor[HTML]{FBEAE1}\textbf{65.54} & \cellcolor[HTML]{FBEAE1}\textbf{54.46} & \cellcolor[HTML]{FBEAE1}\textbf{76.32} & \cellcolor[HTML]{FBEAE1}55.19 & \cellcolor[HTML]{FBEAE1}80.77 & \cellcolor[HTML]{FBEAE1}\textbf{64.56} & \cellcolor[HTML]{FBEAE1}\textbf{66.14}
                             & \cellcolor[HTML]{FBEAE1}\underline{45.04} & \cellcolor[HTML]{FBEAE1}27.82 & \cellcolor[HTML]{FBEAE1}5.89 & \cellcolor[HTML]{FBEAE1}\underline{6.26} & \cellcolor[HTML]{FBEAE1}\underline{16.16} & \cellcolor[HTML]{FBEAE1}\underline{23.34} & \cellcolor[HTML]{FBEAE1}\underline{20.75} \\
          \multirow{-9}{*}{\textbf{Llama3.1-8B}}  &\cellcolor[HTML]{F9E4D9}\ours Iter-4~(Ours) & \cellcolor[HTML]{F9E4D9}\underline{63.23} & \cellcolor[HTML]{F9E4D9}\underline{54.44} & \cellcolor[HTML]{F9E4D9}\underline{76.24} & \cellcolor[HTML]{F9E4D9}\underline{55.34} & \cellcolor[HTML]{F9E4D9}\textbf{80.91} & \cellcolor[HTML]{F9E4D9}64.46 & \cellcolor[HTML]{F9E4D9}\underline{65.77}
                             & \cellcolor[HTML]{F9E4D9}\textbf{45.12} & \cellcolor[HTML]{F9E4D9}28.14 & \cellcolor[HTML]{F9E4D9}\textbf{6.34} & \cellcolor[HTML]{F9E4D9}6.15 & \cellcolor[HTML]{F9E4D9}15.73 & \cellcolor[HTML]{F9E4D9}\textbf{23.56} & \cellcolor[HTML]{F9E4D9}\textbf{20.84} \\
    \bottomrule
    \end{tabular}
    }
    \vspace{-8pt}
    \caption{Performance comparison across Open LLM Leaderboard V1 and V2. \textbf{Bold} indicates best, \underline{underline} indicates second-best (excluding the DPO method). Some datasets are abbreviated, see Appendix A for details.}
    \vspace{-15pt}
    \label{tab:main}
    
\end{table*}

\section{Experiments}
\subsection{Experiment Setup}

\paragraph{Model and Baseline.}

We employ Qwen2.5-1.5B~\citep{yang2024qwen2} and Llama3.1-8B~\citep{dubey2024llama} as our base pretrained models. To ensure a fair comparison under similar training data conditions, we adopt supervised fine-tuning (SFT) and SPIN~\citep{chen2024self} as baseline methods. For SPIN, we follow the official implementation, noting that “SPIN iter-1” refers to the model after the first iteration, which differs slightly from its original definition. Additionally, we perform supplementary experiments using DPO trained on the UltraFeedback dataset as an extended baseline.

\paragraph{Datasets.}
For SFT, we use the full Ultrachat200k dataset, a filtered version of UltraChat~\citep{ding2023enhancing}.
For both \ours and SPIN, we train on a 50k subset of Ultrachat200k. To assess the \ours's performance on domain-specific tasks, we additionally train it using the GSM8K~\citep{cobbe2021gsm8k}.

\paragraph{Evaluation.}

We evaluate models on both versions of the Open LLM Leaderboard (V1~\citep{open-llm-leaderboard-v1} and V2~\citep{open-llm-leaderboard-v2}) using standardized metrics from the Language Model Evaluation Harness~\citep{eval-harness}.
V2 is more challenging than V1, and together they cover 12 benchmarks across knowledge, reasoning, math, and instruction following.
Further experimental setup details are provided in Appendix~\ref{app:expset}.

\captionsetup[subfigure]{justification=centering}

\begin{figure*}[t]
\centering

\begin{subfigure}[t]{0.245\textwidth}
\centering
\includegraphics[width=\textwidth]{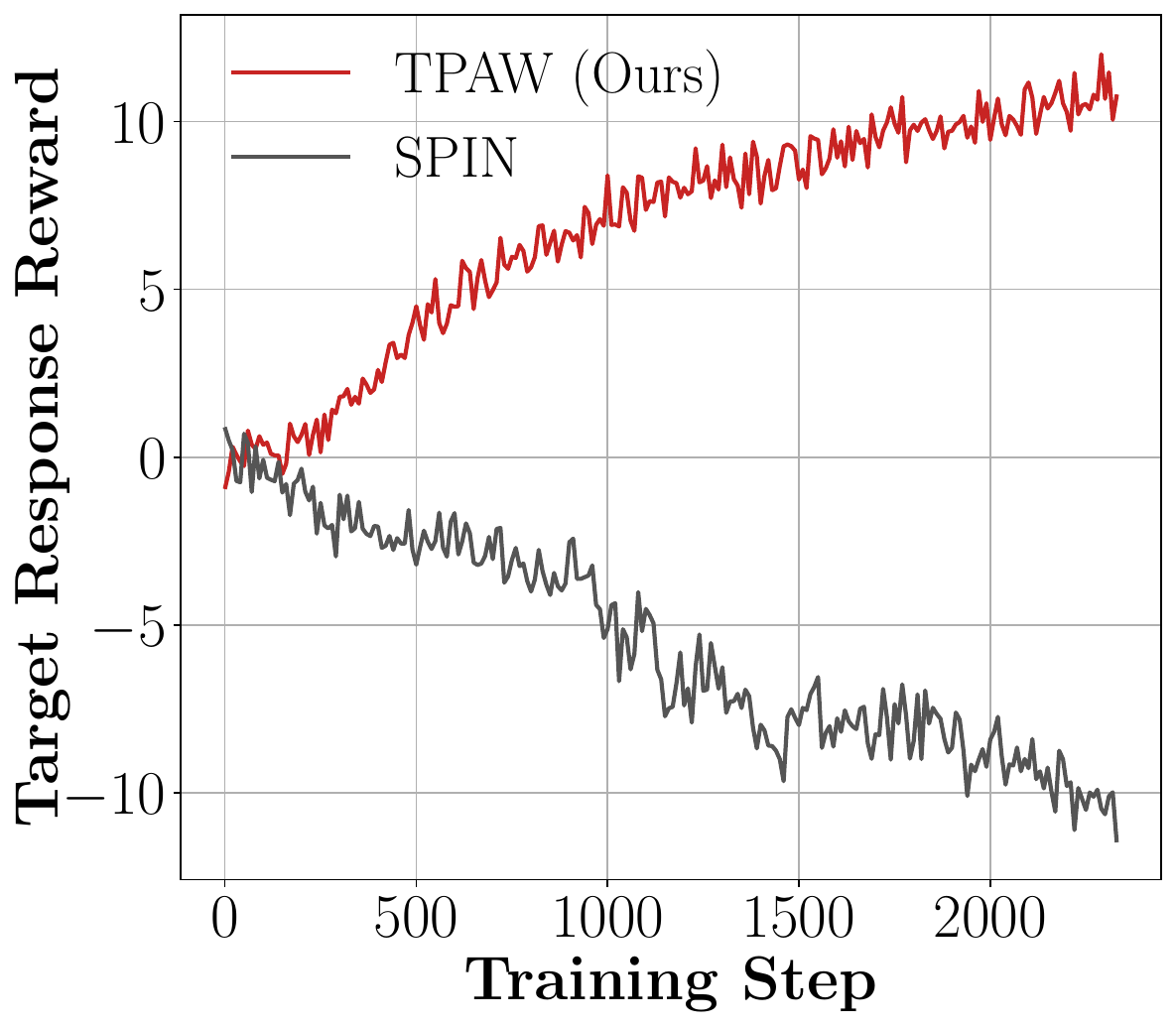}
\caption{Target response reward, trained on Ultrachat200k. \label{fig:reward_chat}}
\end{subfigure}
\begin{subfigure}[t]{0.245\textwidth}
\centering
\includegraphics[width=\textwidth]{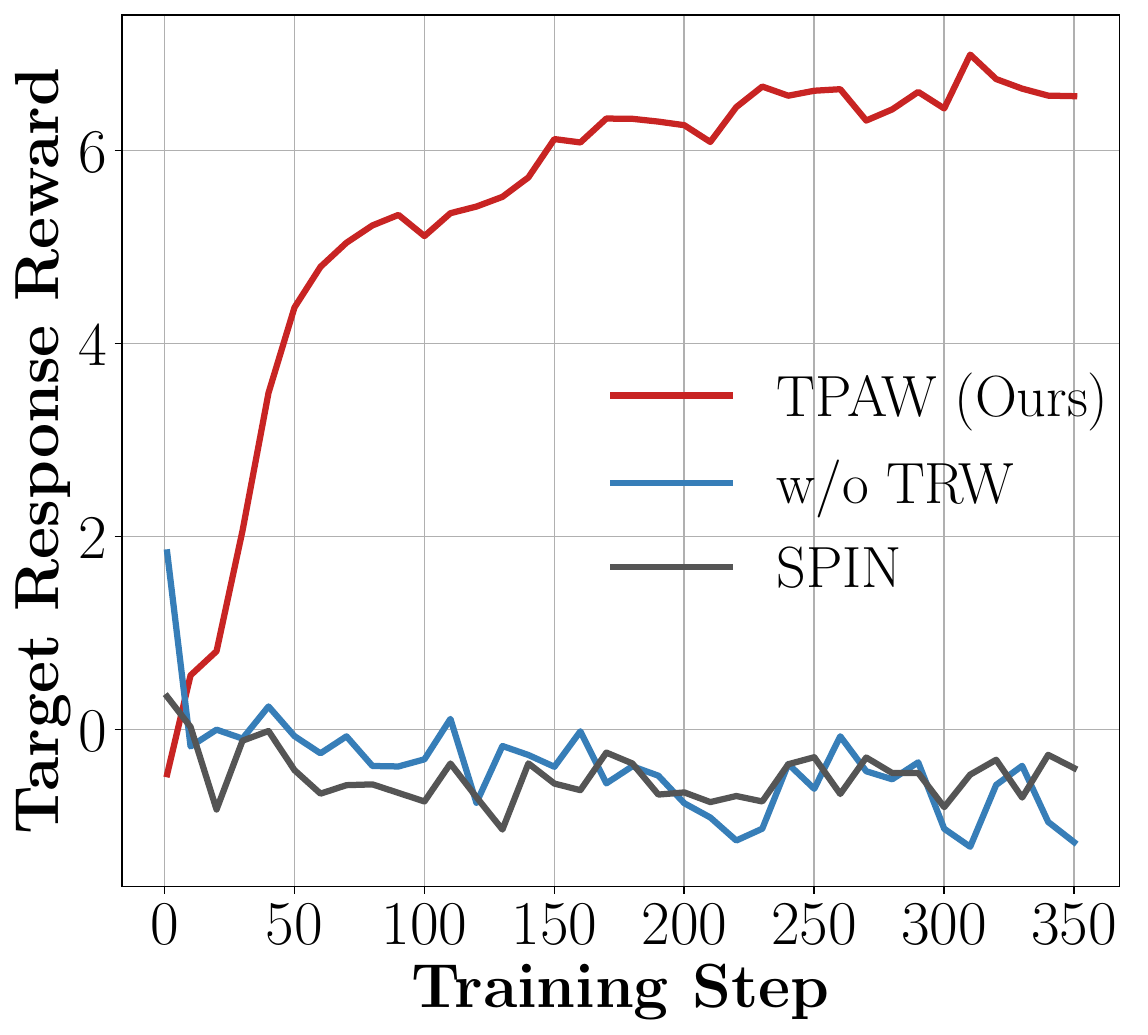}
\caption{Target response reward, trained on GSM8k. \label{fig:reward_gsm}}
\end{subfigure}
\begin{subfigure}[t]{0.245\textwidth}
\centering
\includegraphics[width=\textwidth]{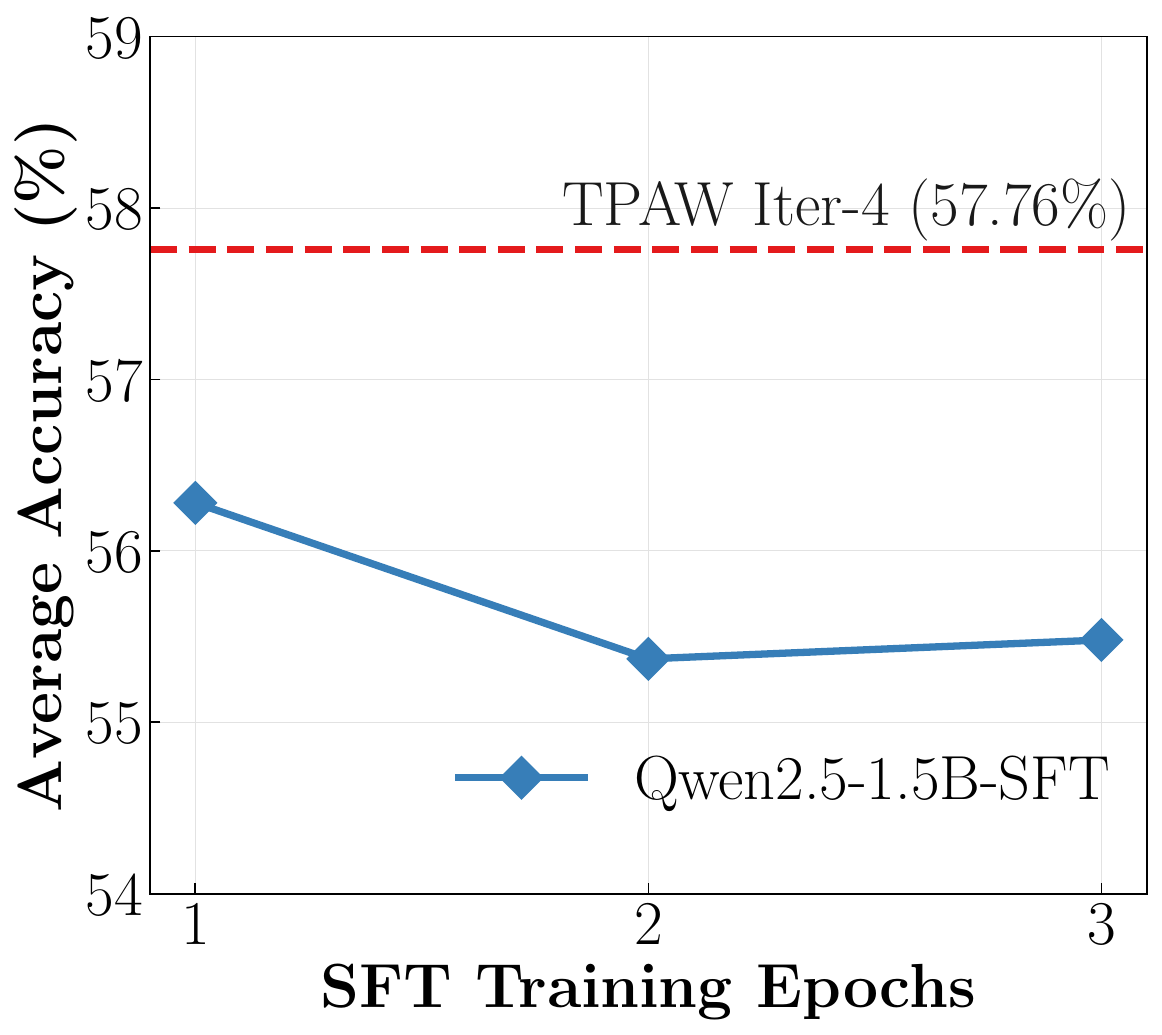}
\caption{Average accuracy of leaderboard v1. \label{fig:sftv1}}
\end{subfigure}
\begin{subfigure}[t]{0.245\textwidth}
\centering
\includegraphics[width=\textwidth]{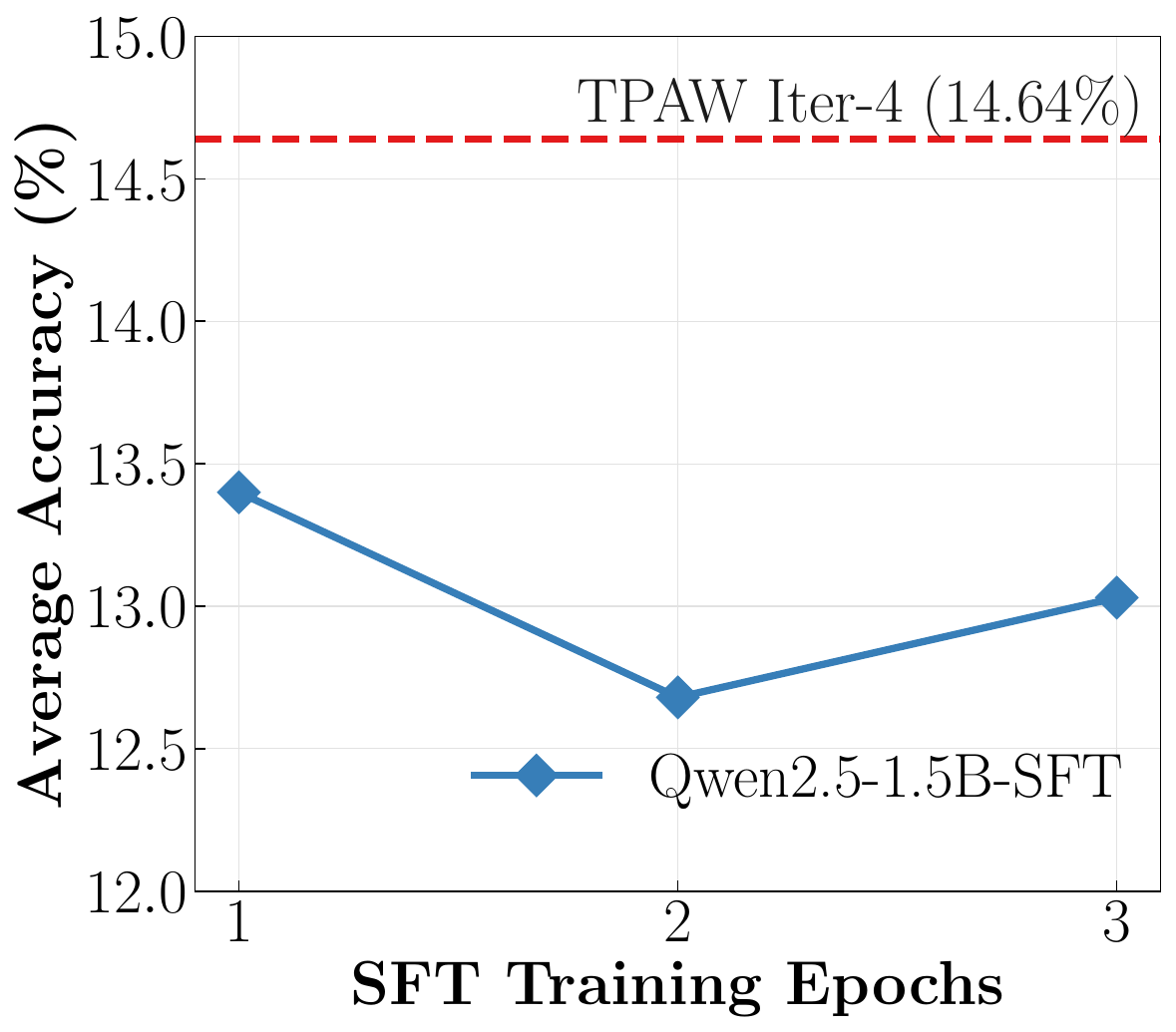}
\caption{Average accuracy of leaderboard v2. \label{fig:sftv2}}
\end{subfigure}
\vspace{-4pt}

\caption{Subfigures (a) and (b) show the target responses reward curves from the iteration 4 training process on Ultrachat200k and GSM8k, respectively. 
Subfigures (c) and (d) present the average accuracy on the leaderboard for models trained with additional epochs using SFT, evaluating the effect of extended training. \label{fig:four_in_row}}
\vspace{-9pt}

\end{figure*}

\subsection{Main Results}

\paragraph{Results on Open LLM Leaderboard.}


Table~\ref{tab:main} reports the evaluation results of \ours on both versions of the Open LLM Leaderboard, compared with the base model and the SPIN baseline over multiple iterations. \ours consistently outperforms both SPIN and the SFT baseline by utilizing the SFT dataset more effectively. Although trained only on a general-purpose dialogue dataset, \ours achieves improvements across all 12 benchmarks, showing strong generalization beyond its training domain. Even with just one-quarter of the original SFT data, it delivers notable gains—for example, on Qwen, up to 4.37\% on IFEval, 3.55\% on Math, and 3.79\% on GSM8k, and on Llama, up to 4.79\% on Arc, 8.78\% on IFEval, and 4.45\% on MUSR. The performance peaks around the third or fourth iteration and then converges. Furthermore, compared with DPO, \ours is trained using only one-quarter of their data volume and does not rely on any additional preference annotations. Despite these more restrictive training conditions, our method still achieves superior performance, particularly on the more challenging Leaderboard V2, highlighting its strong data efficiency and robustness.

These results underscore the effectiveness and robustness of our iterative approach. Building upon the SFT model, we are able to leverage training data more efficiently, producing outputs that align more closely with the target responses—and in some cases, even surpassing previous performance limits—without requiring additional data or external assistance.
The performance boost over SPIN stems from our novel Team-Based Self-Play Framework and Dual Adaptive Weighting Mechanism, which work together to guide learning more efficiently and maintain high-quality outputs throughout the iterative process.


\begin{table}[t]
    \small
    \centering
    \setlength{\tabcolsep}{9pt} 

    {
        \begin{tabular}{c|c}
            \toprule
                Model & Acc~(\%)  \\
            \midrule
                \texttt{Qwen2.5-1.5B-SFT} & 51.25  \\
            \midrule
                \texttt{SPIN}-gsm8k Iter-1 & 53.75 $_{\scalebox{0.85}{(+2.65)}}$   \\
                \texttt{SPIN}-gsm8k Iter-2 & 54.36 $_{\scalebox{0.85}{(+0.61)}}$   \\
                \texttt{SPIN}-gsm8k Iter-3 & 53.75 $_{\scalebox{0.85}{(-0.61)}}$  \\
                \texttt{SPIN}-gsm8k Iter-4 & 54.59 $_{\scalebox{0.85}{(+0.84)}}$   \\
            \midrule
                \ours-gsm8k Iter-1 & 54.21 $_{\scalebox{0.85}{(+3.11)}}$  \\
                \ours-gsm8k Iter-2 & 54.81 $_{\scalebox{0.85}{(+0.60)}}$   \\
                \ours-gsm8k Iter-3 & \underline{56.56} $_{\scalebox{0.85}{(+1.75)}}$   \\
                \ours-gsm8k Iter-4 & \textbf{56.94} $_{\scalebox{0.85}{(+0.38)}}$  \\
            \bottomrule
        \end{tabular}
    }
    \caption{Performance of \ours on GSM8k. Both SPIN and \ours are trained on the GSM8k training set. }
    \label{tab:gsmk}
    \vspace{5pt}
\end{table}

\begin{figure}[t]
    \centering
    \includegraphics[width=0.4\textwidth]{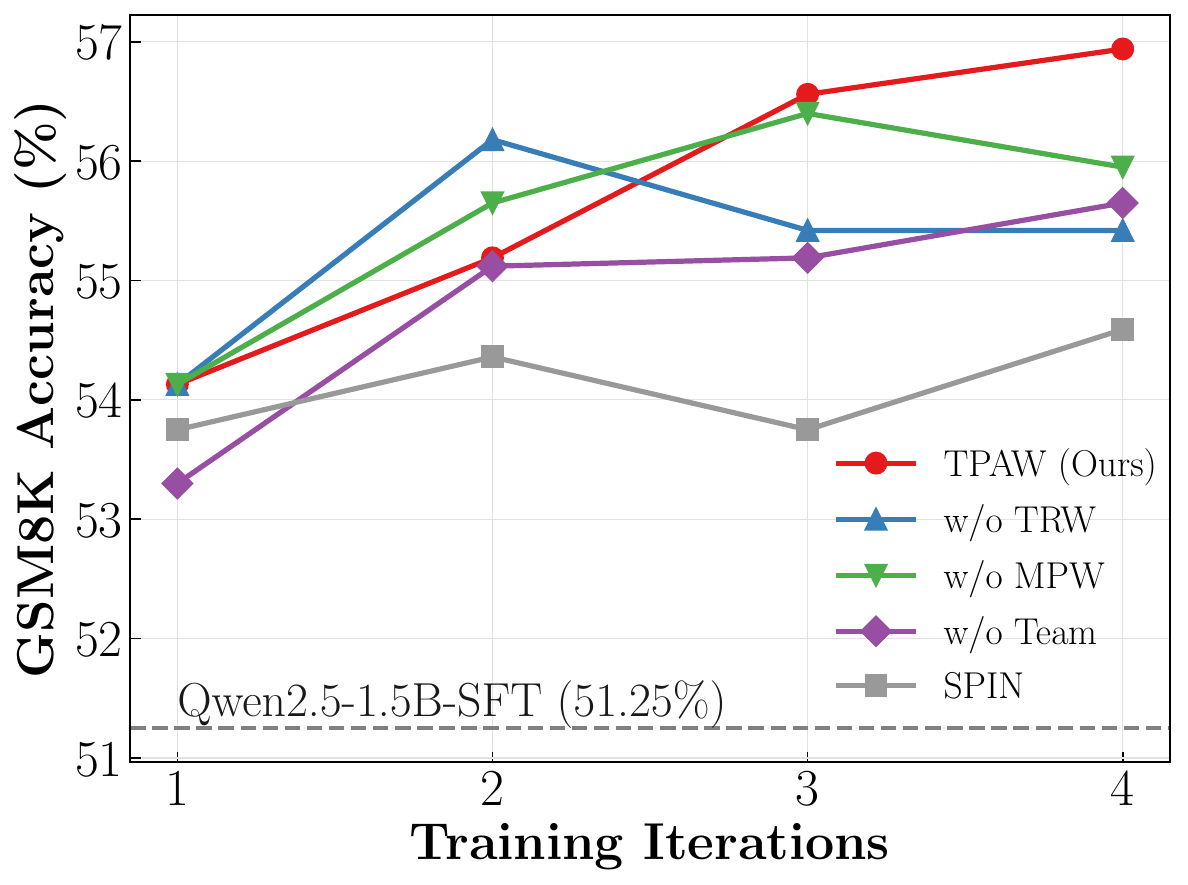}
    \vspace{-3pt}
    \caption{Ablation study on GSM8k. We evaluate \ours by removing key components: without Target Response Weighting (\textbf{w/o TRW}); without Main Player Weighting (\textbf{w/o MPW}); without Team-based Mechanism (\textbf{w/o Team}).}
    \label{fig:ablation}
    \vspace{4pt}
\end{figure}

\paragraph{Results on GSM8K.}
We applied our proposed \ours method for domain-specific fine-tuning on the GSM8K dataset, starting from the Qwen2.5-1.5B-SFT model. Table~\ref{tab:gsmk} compares the accuracy of \ours against the SPIN and standard SFT baselines across multiple iterations.

Both iterative approaches outperform the SFT model, but \ours consistently achieves superior performance at every stage. The method demonstrates steady improvement across iterations, reaching 56.94\% accuracy by the fourth iteration—a 5.69\% gain over the initial SFT model. These results underscore the effectiveness and stability of \ours in enhancing mathematical reasoning capabilities. The greater performance gains observed on domain-specific reasoning and mathematical tasks may stem from a larger initial discrepancy between the model's outputs and the target responses. Compared to the baseline, our method achieves a deeper alignment with the target responses, particularly on such challenging reasoning tasks. Appendix~\ref{app:gsm8k-case} further provides representative GSM8K case studies, showing that these gains mainly come from improved structural reasoning consistency, while token-level arithmetic errors can still remain.

\begin{figure*}[t]
\centering

\begin{subfigure}[t]{0.4\textwidth}
\centering
\includegraphics[width=\textwidth]{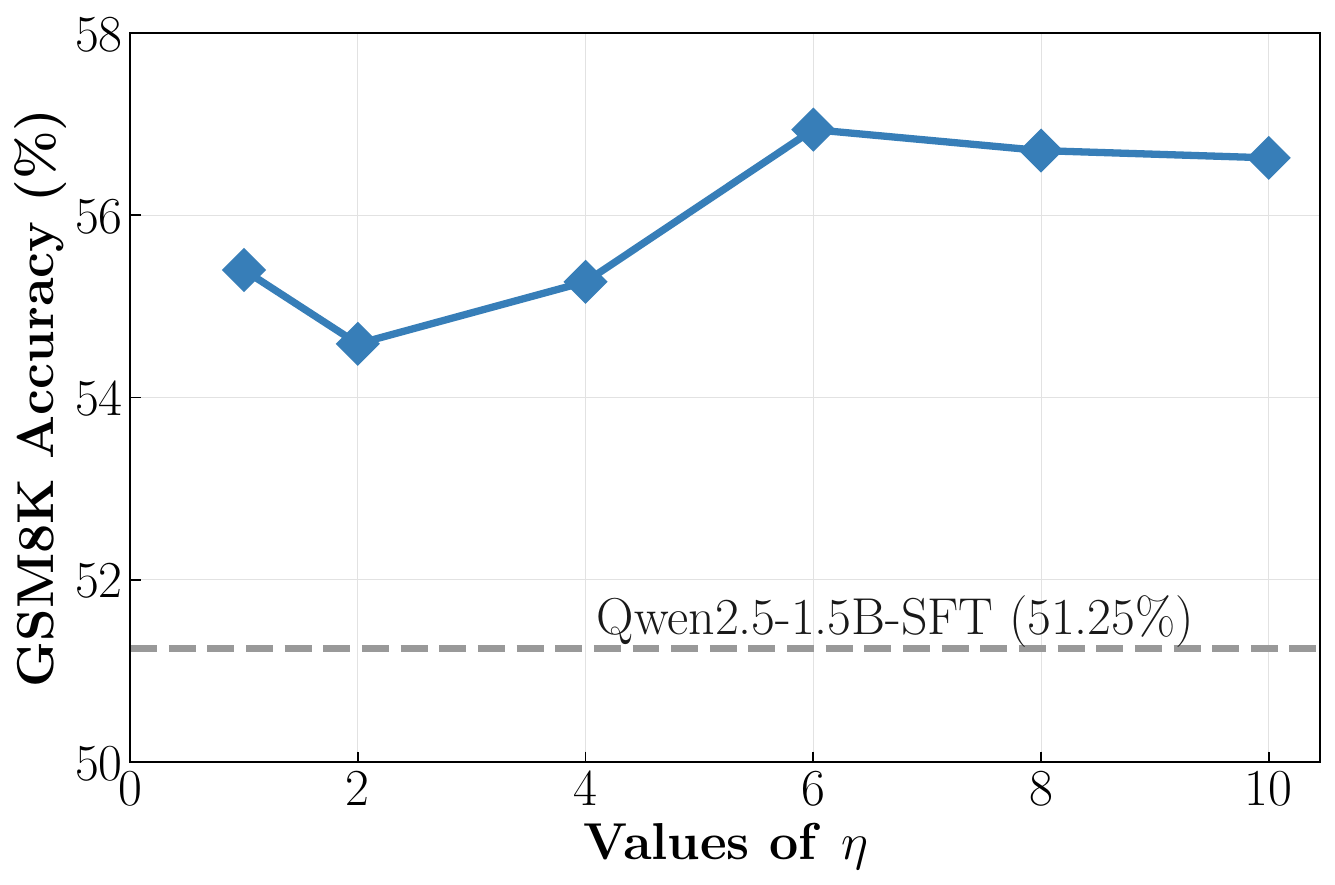}
\end{subfigure}
\hspace{4.5em}
\begin{subfigure}[t]{0.4\textwidth}
\centering
\includegraphics[width=\textwidth]{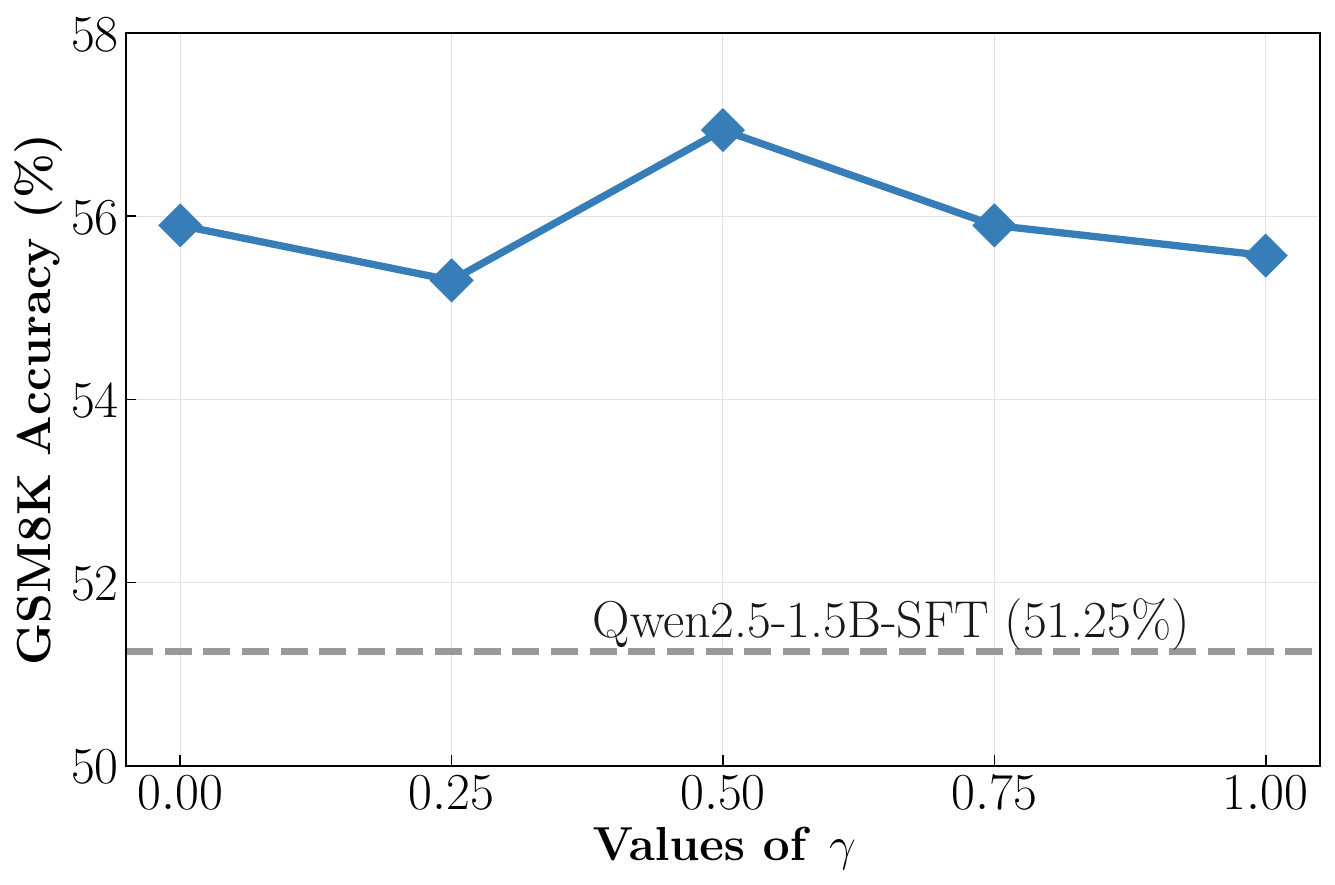}
\end{subfigure}

\vspace{-7pt}

\caption{Impact of hyperparameters on GSM8K accuracy. Performance data is from the fourth iteration.\label{fig:two_in_row}}

\vspace{-11pt}

\end{figure*}

\subsection{Ablation Studies}

To further evaluate the effectiveness of the team-based self-play framework and the dual adaptive weighting mechanism, we conducted additional ablation experiments on GSM8k. In Figure~\ref{fig:ablation}, we present the results of ablating each key component of \ours:

\vspace{-1pt}

\begin{itemize}[left=0cm]
  \setlength{\itemsep}{3pt}  
  \setlength{\parskip}{0pt}  
  \setlength{\parsep}{0pt}   
  \item Without adaptive Target Response Weighting (\textbf{w/o TRW}): Remove adaptive weighting for target responses, with weight fixed at $1$;
  \item Without adaptive Main Player Weighting (\textbf{w/o MPW}): Remove adaptive weighting for main players, all players assigned equal static weights;
  \item Without Team based Mechanism (\textbf{w/o Team}): Remove team structure, using only a single player for both opponent and main player teams.
\end{itemize}

The ablation study results demonstrate that both the team-based self-play framework and the dual adaptive weighting mechanism in \ours are critical to achieving optimal performance: removing either component prevents the model from converging to the optimal solution. Overall, across the four iterations, ablating the team-based mechanism leads to the most significant performance degradation. Notably, even in configurations where these components are removed, our method consistently surpasses baseline performance.
See more ablation experiments and detailed results in Appendix~\ref{app:abalation}.


\subsection{Further Analyses}
\label{sec:analyse}

\paragraph{Adaptive Target Response Weighting Prevents Decline in Target Reward.}
As shown in Figure~\ref{fig:reward_chat} and Figure~\ref{fig:reward_gsm}, training methods that do not employ the Adaptive Target Response Weighting mechanism exhibit a continuous decline in the reward value for the target response, which remains in the negative range. This suggests that during self-play training, the model reduces the output probabilities of both $y^+$ and $y^-$, causing a drift in the output distribution away from the intended target distribution.

In contrast, when our Adaptive Target Response Weighting mechanism is applied, this issue is substantially alleviated. During the training process of \ours, the reward value consistently increases and eventually converges to a positive value, indicating that the model effectively learns to align with the target distribution.

\paragraph{TPAW Makes Fuller Use of Datasets Than More Epochs of SFT.}

As shown in Figure~\ref{fig:sftv1} and Figure~\ref{fig:sftv2}, training for more epochs on the dataset using SFT does not lead to further performance improvements. 
This stems from inherent limitations of the SFT training objective, causing the model to overfit after multiple epochs by mechanically memorizing answers in the dataset rather than genuinely learning generalizable capabilities.
In contrast, by continuing training on the same dataset, \ours successfully surpasses the performance ceiling of SFT, enabling the model not only to fit the target data distribution but also to retain its generalization ability. See more detail results in Appendix~\ref{app:moresft}.

\paragraph{Analysis and Ablation on Hyperparameters $\bm{\eta}$ and $\bm{\gamma}$.} 
Figure~\ref{fig:two_in_row} presents the results of a numerical analysis investigating their impact.
$\bm{\eta}$ and $\bm{\gamma}$ are core components of \ours's Dual Adaptive Weighting mechanism, with their practical values set to 6 and 0.5.
$\bm{\eta}$ controls how strongly the weight increases when the target response reward is low. 
The results show that relatively higher values of $\bm{\eta}$ encourage greater alignment with the target response, resulting in improved performance.
$\bm{\gamma}$ adjusts the sensitivity of the main players’ weight distribution to their judgment ability. A larger $\bm{\gamma}$ amplifies the impact of weaker performance, assigning more weight to poorly performing plyers during optimization. In contrast, when $\bm{\gamma}$ is close to zero, weights become nearly uniform, ignoring differences among players.
The results show that a moderate value of 0.5 yields the most balanced weight distribution.
See more experiments and detailed result in Appendix~\ref{app:hyper}.

\section{Conclusion}
In this work, we introduced \ours, a novel self-training algorithm that enhances LLM alignment through a team-based self-play framework augmented with dual adaptive weighting mechanisms. \ours leverages past policy checkpoints both as opponents and main players, enabling more robust and stable policy optimization.
By adaptively adjusting the influence of individual team members and reweighting target responses, \ours effectively mitigates issues related to noisy synthetic data and output offset from target response distribution.
Extensive empirical results demonstrate that \ours significantly improves model alignment and generalization across various benchmarks, validating its ability to better exploit SFT datasets and align with the target response distribution. 
We believe our team-based, adaptively weighted self-play paradigm paves the way for scalable and effective LLM alignment with minimal human supervision.

\section*{Acknowledgements}

This work was supported in part by National Key R\&D Program of China (SQ2024YFE0200592), National Natural Science Foundation of China (62476070), Shenzhen Science and Technology Program \seqsplit{(JCYJ20241202123503005, \, GXWD20231128103232001, \, ZDSYS20230626091203008, \, KQTD20240729102154066)}, Department of Science and Technology of Guangdong (2024A1515011540) and Suzhou Science and Technology Program (SYG2025072).

\section*{Limitations}

While our work demonstrates the effectiveness of \ours in aligning model outputs to target distributions, its performance is inherently bounded by the quality of the SFT datasets. This highlights both a limitation and a potential opportunity for future research in model distillation task. Additionally, our experiments on GSM8K suggest that \ours holds promise for domain-specific fine-tuning, highlighting the need for further fine-tuning and evaluation across a broader range of specialized tasks. Overall, addressing these limitations will be crucial for realizing the full potential of our approach in broader and more challenging settings. In addition, our ablations suggest a trade-off in team size $N_{\max}$: while a larger team can improve diversity, setting $N_{\max}>3$ may introduce weaker early checkpoints and lower-quality preference signals, leading to slight performance degradation.

\bibliography{custom}

@inproceedings{chen2024self,
  title={Self-play fine-tuning convertsweak language models to strong language models},
  author={Chen, Zixiang and Deng, Yihe and Yuan, Huizhuo and Ji, Kaixuan and Gu, Quanquan},
  booktitle={Proceedings of the 41st International Conference on Machine Learning (ICML)},
  pages={6621--6642},
  year={2024}
}

@article{rafailov2023direct,
  title={Direct preference optimization: Your language model is secretly a reward model},
  author={Rafailov, Rafael and Sharma, Archit and Mitchell, Eric and Manning, Christopher D and Ermon, Stefano and Finn, Chelsea},
  journal={Advances in Neural Information Processing Systems (NeurIPS)},
  volume={36},
  pages={53728--53741},
  year={2023}
}

@article{ouyang2022training,
  title={Training language models to follow instructions with human feedback},
  author={Ouyang, Long and Wu, Jeffrey and Jiang, Xu and Almeida, Diogo and Wainwright, Carroll and Mishkin, Pamela and Zhang, Chong and Agarwal, Sandhini and Slama, Katarina and Ray, Alex and others},
  journal={Advances in neural information processing systems (NeurIPS)},
  volume={35},
  pages={27730--27744},
  year={2022}
}

@article{christiano2017deep,
  title={Deep reinforcement learning from human preferences},
  author={Christiano, Paul F and Leike, Jan and Brown, Tom and Martic, Miljan and Legg, Shane and Amodei, Dario},
  journal={Advances in neural information processing systems (NeurIPS)},
  volume={30},
  year={2017}
}

@article{stiennon2020learning,
  title={Learning to summarize with human feedback},
  author={Stiennon, Nisan and Ouyang, Long and Wu, Jeffrey and Ziegler, Daniel and Lowe, Ryan and Voss, Chelsea and Radford, Alec and Amodei, Dario and Christiano, Paul F},
  journal={Advances in neural information processing systems (NeurIPS)},
  volume={33},
  pages={3008--3021},
  year={2020}
}

@article{radford2019language,
  title={Language models are unsupervised multitask learners},
  author={Radford, Alec and Wu, Jeffrey and Child, Rewon and Luan, David and Amodei, Dario and Sutskever, Ilya and others},
  journal={OpenAI blog},
  volume={1},
  number={8},
  pages={9},
  year={2019}
}

@article{achiam2023gpt,
  title={Gpt-4 technical report},
  author={Achiam, Josh and Adler, Steven and Agarwal, Sandhini and Ahmad, Lama and Akkaya, Ilge and Aleman, Florencia Leoni and Almeida, Diogo and Altenschmidt, Janko and Altman, Sam and Anadkat, Shyamal and others},
  journal={arXiv preprint arXiv:2303.08774},
  year={2023}
}

@inproceedings{DBLP:conf/icml/YuanPCLSXW24,
  author       = {Weizhe Yuan and
                  Richard Yuanzhe Pang and
                  Kyunghyun Cho and
                  Xian Li and
                  Sainbayar Sukhbaatar and
                  Jing Xu and
                  Jason Weston},
  title        = {Self-Rewarding Language Models},
  booktitle    = {Forty-first International Conference on Machine Learning (ICML)},
  year         = {2024},
}

@inproceedings{DBLP:conf/acl/WangKMLSKH23,
  author       = {Yizhong Wang and
                  Yeganeh Kordi and
                  Swaroop Mishra and
                  Alisa Liu and
                  Noah A. Smith and
                  Daniel Khashabi and
                  Hannaneh Hajishirzi},
  editor       = {Anna Rogers and
                  Jordan L. Boyd{-}Graber and
                  Naoaki Okazaki},
  title        = {Self-Instruct: Aligning Language Models with Self-Generated Instructions},
  booktitle    = {Proceedings of the 61st Annual Meeting of the Association for Computational
                  Linguistics (ACL)},
  pages        = {13484--13508},
  year         = {2023},
}

@inproceedings{
wang2025cream,
title={{CREAM}: Consistency Regularized Self-Rewarding Language Models},
author={Zhaoyang Wang and Weilei He and Zhiyuan Liang and Xuchao Zhang and Chetan Bansal and Ying Wei and Weitong Zhang and Huaxiu Yao},
booktitle={The Thirteenth International Conference on Learning Representations (ICLR)},
year={2025},
}

@article{pal2024smaug,
  title={Smaug: Fixing failure modes of preference optimisation with dpo-positive},
  author={Pal, Arka and Karkhanis, Deep and Dooley, Samuel and Roberts, Manley and Naidu, Siddartha and White, Colin},
  journal={arXiv preprint arXiv:2402.13228},
  year={2024}
}

@inproceedings{
razin2025unintentional,
title={Unintentional Unalignment: Likelihood Displacement in Direct Preference Optimization},
author={Noam Razin and Sadhika Malladi and Adithya Bhaskar and Danqi Chen and Sanjeev Arora and Boris Hanin},
booktitle={The Thirteenth International Conference on Learning Representations (ICLR)},
year={2025},
}

@inproceedings{
ren2025learning,
title={Learning Dynamics of {LLM} Finetuning},
author={Yi Ren and Danica J. Sutherland},
booktitle={The Thirteenth International Conference on Learning Representations (ICLR)},
year={2025},
}

@inproceedings{
ren2024bias,
title={Bias Amplification in Language Model Evolution: An Iterated Learning Perspective},
author={Yi Ren and Shangmin Guo and Linlu Qiu and Bailin Wang and Danica J. Sutherland},
booktitle={The Thirty-eighth Annual Conference on Neural Information Processing Systems (NeurIPS)},
year={2024},
}

@inproceedings{xu2024pride,
  title={Pride and Prejudice: LLM Amplifies Self-Bias in Self-Refinement},
  author={Xu, Wenda and Zhu, Guanglei and Zhao, Xuandong and Pan, Liangming and Li, Lei and Wang, William},
  booktitle={Proceedings of the 62nd Annual Meeting of the Association for Computational Linguistics (ACL)},
  pages={15474--15492},
  year={2024}
}

@inproceedings{
huang2024self,
title={Self-Improvement in Language Models: The Sharpening Mechanism},
author={Audrey Huang and Adam Block and Dylan J Foster and Dhruv Rohatgi and Cyril Zhang and Max Simchowitz and Jordan T. Ash and Akshay Krishnamurthy},
booktitle={The Thirteenth International Conference on Learning Representations (ICLR)},
year={2025},
}

@inproceedings{
song2025mind,
title={Mind the Gap: Examining the Self-Improvement Capabilities of Large Language Models},
author={Yuda Song and Hanlin Zhang and Carson Eisenach and Sham M. Kakade and Dean Foster and Udaya Ghai},
booktitle={The Thirteenth International Conference on Learning Representations (ICLR)},
year={2025},
}

@inproceedings{
yuan2025advancing,
title={Advancing {LLM} Reasoning Generalists with Preference Trees},
author={Lifan Yuan and Ganqu Cui and Hanbin Wang and Ning Ding and Xingyao Wang and Boji Shan and Zeyuan Liu and Jia Deng and Huimin Chen and Ruobing Xie and Yankai Lin and Zhenghao Liu and Bowen Zhou and Hao Peng and Zhiyuan Liu and Maosong Sun},
booktitle={The Thirteenth International Conference on Learning Representations (ICLR)},
year={2025},
}

@inproceedings{
rafailov2024from,
title={From \$r\$ to \$Q{\textasciicircum}*\$: Your Language Model is Secretly a Q-Function},
author={Rafael Rafailov and Joey Hejna and Ryan Park and Chelsea Finn},
booktitle={First Conference on Language Modeling},
year={2024},
}

@inproceedings{tajwar2024preference,
  title={Preference fine-tuning of LLMs should leverage suboptimal, on-policy data},
  author={Tajwar, Fahim and Singh, Anikait and Sharma, Archit and Rafailov, Rafael and Schneider, Jeff and Xie, Tengyang and Ermon, Stefano and Finn, Chelsea and Kumar, Aviral},
  booktitle={Proceedings of the 41st International Conference on Machine Learning (ICML)},
  pages={47441--47474},
  year={2024}
}

@article{pang2024iterative,
  title={Iterative reasoning preference optimization},
  author={Pang, Richard Yuanzhe and Yuan, Weizhe and He, He and Cho, Kyunghyun and Sukhbaatar, Sainbayar and Weston, Jason},
  journal={Advances in Neural Information Processing Systems (NeurIPS)},
  volume={37},
  pages={116617--116637},
  year={2024}
}

@article{yang2024qwen2,
  title={Qwen2. 5 technical report},
  author={Yang, An and Yang, Baosong and Zhang, Beichen and Hui, Binyuan and Zheng, Bo and Yu, Bowen and Li, Chengyuan and Liu, Dayiheng and Huang, Fei and Wei, Haoran and others},
  journal={arXiv preprint arXiv:2412.15115},
  year={2024}
}

@article{dubey2024llama,
  title={The llama 3 herd of models},
  author={Dubey, Abhimanyu and Jauhri, Abhinav and Pandey, Abhinav and Kadian, Abhishek and Al-Dahle, Ahmad and Letman, Aiesha and Mathur, Akhil and Schelten, Alan and Yang, Amy and Fan, Angela and others},
  journal={arXiv e-prints},
  pages={arXiv--2407},
  year={2024}
}

@inproceedings{ding2023enhancing,
  title={Enhancing Chat Language Models by Scaling High-quality Instructional Conversations},
  author={Ding, Ning and Chen, Yulin and Xu, Bokai and Qin, Yujia and Hu, Shengding and Liu, Zhiyuan and Sun, Maosong and Zhou, Bowen},
  booktitle={Proceedings of the 2023 Conference on Empirical Methods in Natural Language Processing (EMNLP)},
  pages={3029--3051},
  year={2023}
}

@article{cobbe2021gsm8k,
  title={Training Verifiers to Solve Math Word Problems},
  author={Cobbe, Karl and Kosaraju, Vineet and Bavarian, Mohammad and Chen, Mark and Jun, Heewoo and Kaiser, Lukasz and Plappert, Matthias and Tworek, Jerry and Hilton, Jacob and Nakano, Reiichiro and Hesse, Christopher and Schulman, John},
  journal={arXiv preprint arXiv:2110.14168},
  year={2021}
}

@misc{open-llm-leaderboard-v2,
  author = {Clémentine Fourrier and Nathan Habib and Alina Lozovskaya and Konrad Szafer and Thomas Wolf},
  title = {Open LLM Leaderboard v2},
  year = {2024},
  publisher = {Hugging Face},
  howpublished = "\url{https://huggingface.co/spaces/open-llm-leaderboard/open_llm_leaderboard}",
}

@misc{eval-harness,
  author       = {Gao, Leo and Tow, Jonathan and Abbasi, Baber and Biderman, Stella and Black, Sid and DiPofi, Anthony and Foster, Charles and Golding, Laurence and Hsu, Jeffrey and Le Noac'h, Alain and Li, Haonan and McDonell, Kyle and Muennighoff, Niklas and Ociepa, Chris and Phang, Jason and Reynolds, Laria and Schoelkopf, Hailey and Skowron, Aviya and Sutawika, Lintang and Tang, Eric and Thite, Anish and Wang, Ben and Wang, Kevin and Zou, Andy},
  title        = {The Language Model Evaluation Harness},
  month        = 07,
  year         = 2024,
  publisher    = {Zenodo},
  version      = {v0.4.3},
  doi          = {10.5281/zenodo.12608602},
  url          = {https://zenodo.org/records/12608602}
}

@misc{open-llm-leaderboard-v1,
  author = {Edward Beeching and Clémentine Fourrier and Nathan Habib and Sheon Han and Nathan Lambert and Nazneen Rajani and Omar Sanseviero and Lewis Tunstall and Thomas Wolf},
  title = {Open LLM Leaderboard (2023-2024)},
  year = {2023},
  publisher = {Hugging Face},
  howpublished = "\url{https://huggingface.co/spaces/open-llm-leaderboard-old/open_llm_leaderboard}"
}

@article{silver2017mastering,
  title={Mastering the game of go without human knowledge},
  author={Silver, David and Schrittwieser, Julian and Simonyan, Karen and Antonoglou, Ioannis and Huang, Aja and Guez, Arthur and Hubert, Thomas and Baker, Lucas and Lai, Matthew and Bolton, Adrian and others},
  journal={nature},
  volume={550},
  number={7676},
  pages={354--359},
  year={2017},
}

@article{dong2024self,
  title={Self-play with execution feedback: Improving instruction-following capabilities of large language models},
  author={Dong, Guanting and Lu, Keming and Li, Chengpeng and Xia, Tingyu and Yu, Bowen and Zhou, Chang and Zhou, Jingren},
  journal={arXiv preprint arXiv:2406.13542},
  year={2024}
}

@inproceedings{DBLP:conf/icml/0001PMMFLBHCRP24,
  author       = {Harrison Lee and
                  Samrat Phatale and
                  Hassan Mansoor and
                  Thomas Mesnard and
                  Johan Ferret and
                  Kellie Lu and
                  Colton Bishop and
                  Ethan Hall and
                  Victor Carbune and
                  Abhinav Rastogi and
                  Sushant Prakash},
  title        = {{RLAIF} vs. {RLHF:} Scaling Reinforcement Learning from Human Feedback
                  with {AI} Feedback},
  booktitle    = {Forty-first International Conference on Machine Learning (ICML))},
  year         = {2024},
}

@article{DBLP:journals/corr/abs-2402-01306,
  title={Kto: Model alignment as prospect theoretic optimization},
  author={Ethayarajh, Kawin and Xu, Winnie and Muennighoff, Niklas and Jurafsky, Dan and Kiela, Douwe},
  journal={arXiv preprint arXiv:2402.01306},
  year={2024}
}

@inproceedings{DBLP:conf/nips/0001X024,
  author       = {Yu Meng and
                  Mengzhou Xia and
                  Danqi Chen},
  title        = {SimPO: Simple Preference Optimization with a Reference-Free Reward},
  booktitle    = {Advances in Neural Information Processing Systems 38: Annual Conference
                  on Neural Information Processing Systems (NeurIPS)},
  year         = {2024},
}

@article{kingma2014adam,
  title={Adam: A method for stochastic optimization},
  author={Kingma, Diederik P},
  journal={arXiv preprint arXiv:1412.6980},
  year={2014}
}

@article{hinton2012neural,
  title={Neural networks for machine learning lecture 6a overview of mini-batch gradient descent},
  author={Hinton, Geoffrey and Srivastava, Nitish and Swersky, Kevin},
  journal={Cited on},
  volume={14},
  number={8},
  pages={2},
  year={2012}
}

@article{dao2023flashattention,
  title={Flashattention-2: Faster attention with better parallelism and work partitioning},
  author={Dao, Tri},
  journal={arXiv preprint arXiv:2307.08691},
  year={2023}
}

@inproceedings{rajbhandari2020zero,
  title={Zero: Memory optimizations toward training trillion parameter models},
  author={Rajbhandari, Samyam and Rasley, Jeff and Ruwase, Olatunji and He, Yuxiong},
  booktitle={SC20: International Conference for High Performance Computing, Networking, Storage and Analysis},
  pages={1--16},
  year={2020},
  organization={IEEE}
}

@article{wu2024self,
  title={Self-play preference optimization for language model alignment},
  author={Wu, Yue and Sun, Zhiqing and Yuan, Huizhuo and Ji, Kaixuan and Yang, Yiming and Gu, Quanquan},
  journal={arXiv preprint arXiv:2405.00675},
  year={2024}
}

@article{zhao2025absolute,
  title={Absolute Zero: Reinforced Self-play Reasoning with Zero Data},
  author={Zhao, Andrew and Wu, Yiran and Yue, Yang and Wu, Tong and Xu, Quentin and Lin, Matthieu and Wang, Shenzhi and Wu, Qingyun and Zheng, Zilong and Huang, Gao},
  journal={arXiv preprint arXiv:2505.03335},
  year={2025}
}

@article{cheng2024self,
  title={Self-playing adversarial language game enhances llm reasoning},
  author={Cheng, Pengyu and Dai, Yong and Hu, Tianhao and Xu, Han and Zhang, Zhisong and Han, Lei and Du, Nan and Li, Xiaolong},
  journal={Advances in Neural Information Processing Systems (NeurIPS)},
  volume={37},
  pages={126515--126543},
  year={2024}
}

@article{swamy2024minimaximalist,
  title={A minimaximalist approach to reinforcement learning from human feedback},
  author={Swamy, Gokul and Dann, Christoph and Kidambi, Rahul and Wu, Zhiwei Steven and Agarwal, Alekh},
  journal={arXiv preprint arXiv:2401.04056},
  year={2024}
}

@article{samuel1959some,
  title={Some studies in machine learning using the game of checkers},
  author={Samuel, Arthur L},
  journal={IBM Journal of research and development},
  volume={3},
  number={3},
  pages={210--229},
  year={1959},
  publisher={IBM}
}

@article{tesauro1995temporal,
  title={Temporal difference learning and TD-Gammon},
  author={Tesauro, Gerald and others},
  journal={Communications of the ACM},
  volume={38},
  number={3},
  pages={58--68},
  year={1995}
}

@inproceedings{shi2025safety,
  title={Safety Alignment via Constrained Knowledge Unlearning},
  author={Shi, Zesheng and Zhou, Yigeng and Li, Jing},
  booktitle={Proceedings of the 63rd Annual Meeting of the Association for Computational Linguistics (ACL)},
  year={2025}
}

\appendix

\section{Further Details about Experiment Set-Up}
\label{app:expset}

\subsection{Hyperparameters and Implementation Details}

We employ DeepSpeed ZeRO-3~\citep{rajbhandari2020zero} and FlashAttention-2~\citep{dao2023flashattention} across all training stages to improve efficiency and reduce computational cost.
Qwen2.5-1.5B-SFT and Llama3.1-8B-SFT were trained on the UltraChat200k dataset via LLaMA-Factory, with the following shared settings: batch size of 64, maximum sequence length of 2048, a cosine learning rate schedule (10\% warmup over 1 epoch), and the Adam optimizer \citep{kingma2014adam}. Their distinct learning rates are 5e-5 (for Qwen) and 1e-5 (for Llama).
These SFT models serve as the starting point for all subsequent baseline training, including DPO, SPIN, and \ours.

For DPO training, we conducted additional experiments on the UltraFeedback dataset as a supplementary baseline. Specifically, we used LLaMA-Factory with a learning rate of 1e-6 for 2 epochs.

For \ours, we use a 50k subset of UltraChat200k and adopt the same batch size and sequence length. The model is trained for 2 epochs using RMSProp~\citep{hinton2012neural} optimizer, a cosine learning rate schedule with 10\% warmup, and a regularization coefficient $\bm{\lambda}=0.1$. As UltraChat200k contains multi-round conversations, we only retain the first round as prompt-completion pairs. For the newly introduced hyperparameters in this method, we set $\bm{\gamma}=0.5$, $\bm{\eta}=6$.

During the four-stage training of \ours for Qwen, the learning rate was set to 1e-6 for iterations 1–2, then decayed to 1e-7 for iterations 3–4 as the model converged. For Llama, the learning rate was also 1e-6 in iterations 1–2, but decayed to 5e-7 in iterations 3–4 during convergence.
In iteration 4, we further reduce the training to 1 epoch and increase $\bm{\lambda}$ to 5.0. 
For \ours on GSM8k, we follow the same setup, except that iteration 3 is trained for only 1 epoch.


\begin{table}[t]
    \centering
    \begin{tabular}{c|cccc}
        \toprule
        \textbf{Method} & Iter-1 & Iter-2 & Iter-3 & Iter-4\\
        \midrule
        SPIN & 6:13 & 14:42 & 14:42 & 14:42 \\
        TPAW & 3:20 &  7:02 & 6:10 & 6:10  \\
        \bottomrule
    \end{tabular}
     \vspace{-3pt}
    \caption{Total training time per iteration (hours:minutes). Compared to SPIN, TPAW's Iter3 and 4 were trained for only one epoch. }
    \label{tab:computation}
\end{table}

\subsection{Computational cost analysis}
We analyze the computational and memory overhead of \ours and clarify its efficiency relative to existing self-play methods.

Importantly, only the current policy model is updated via backpropagation during training. All historical checkpoints involved in the main-player team are used exclusively for forward inference, incurring no additional gradient-related memory cost compared to standard fine-tuning.

To further reduce runtime and memory usage, \ours adopts a log-probability precomputation strategy: log-probabilities of all reference models are computed once on the training set prior to training, after which the reference models are fully offloaded from GPU memory. During training, each step computes log-probabilities only for the current policy model. On 4×L40S GPUs, precomputing 50k samples per reference model requires approximately 30 minutes, which is negligible relative to total training time.

In practice, this design leads to higher training throughput. We observe an increase in train\_steps\_per\_second from 0.021 (SPIN) to 0.044 (\ours), corresponding to a 110\% speedup per step. Moreover, even after accounting for precomputation, \ours consistently achieves shorter wall-clock training time across all iterations. 
Table~\ref{tab:computation} reports total training time per iteration on 8×A100 80GB GPUs with Llama-3.1-8B.

Overall, \ours’s forward-only use of historical checkpoints combined with log-probability precomputation ensures a lower effective memory footprint and improved training efficiency, demonstrating superior computational performance compared to prior self-play approaches.

\subsection{Evaluation Datasets}
\subsubsection{Open LLM Leaderboard V1.}

Leaderboard v1 evaluates models on 6 key benchmarks using the Eleuther AI Language Model Evaluation Harness, a unified framework designed to test generative language models on a wide range of different evaluation tasks.

\begin{itemize}\setlength{\itemsep}{0pt}
    \item \textbf{AI2 Reasoning Challenge (25-shot)}: A set of grade-school science questions. Abbreviated as Arc in Table~\ref{tab:main}.
    \item \textbf{HellaSwag (10-shot)}: A commonsense reasoning test, which is easy for humans (~95\% accuracy) but challenging for state-of-the-art models. Abbreviated as HS in Table~\ref{tab:main}.
    \item \textbf{MMLU (5-shot)}: A test to measure a text model's multitask accuracy. The test covers 57 tasks, including elementary mathematics, U.S. history, computer science, law, and more.
    \item \textbf{TruthfulQA (0-shot)}: A test to measure a model's propensity to reproduce falsehoods commonly found online. Note: TruthfulQA is technically a 6-shot task in the Harness because each example is prepended with 6 Q/A pairs, even in the 0-shot setting. Abbreviated as TQA in Table~\ref{tab:main}.
    \item \textbf{Winogrande (5-shot)}: An adversarial and difficult Winograd benchmark at scale, for commonsense reasoning. Abbreviated as Wino in Table~\ref{tab:main}.
    \item \textbf{GSM8k (5-shot)}: Diverse grade school math word problems to measure a model's ability to solve multi-step mathematical reasoning problems.
\end{itemize}

\subsubsection{Open LLM Leaderboard V2.}

Leaderboard v2 seeks new benchmarks using high-quality, uncontaminated datasets and evaluates models on key tasks. They chose the following tasks: knowledge testing, reasoning abilities, complex mathematical abilities, and tasks related to human preferences, such as instruction following. They cover these tasks with six benchmarks:

\begin{itemize} \setlength{\itemsep}{0pt}
    \item \textbf{MMLU-Pro}: A refined version of the MMLU dataset with higher difficulty, with expert-reviewed questions to reduce noise. Abbreviated as MMLU-p in Table~\ref{tab:main}.
    \item \textbf{GPQA}: An extremely challenging knowledge dataset, designed by domain experts to ensure difficulty and factual accuracy.
    \item \textbf{MuSR}: A dataset with complex reasoning problems, such as murder mysteries and team allocation issues, requiring long-range reasoning.
    \item \textbf{MATH}: A compilation of high school-level competition problems, requiring precise output formatting, focusing on the hardest questions.
    \item \textbf{IFEval}: A test of a model's ability to strictly follow instructions, focusing on format execution.
    \item \textbf{BBH}: A subset of 23 challenging tasks from the BigBench dataset, involving multi-step reasoning and commonsense knowledge, with performance highly correlated with human preferences.
\end{itemize}

\subsection{Baselines}

\subsubsection{SFT}

Supervised Fine-Tuning (SFT) is a widely used technique to adapt pre-trained language models (LMs) to specific downstream tasks by fine-tuning on labeled datasets. The goal of SFT is to reduce task-related errors and align the model's predictions with human-labeled ground truth.

Given a pre-trained model \( f_\theta \), the SFT process involves the following:

\textbf{Pre-trained Model.}
The model \( f_\theta(x) \) is initially pre-trained on a large corpus using language modeling objectives, allowing the model to learn useful language representations.

\textbf{Fine-Tuning on Task-Specific Data.}
The pre-trained model is fine-tuned on a supervised dataset \( D = \{(x_i, y_i)\}_{i=1}^N \), where \( x_i \) is the input and \( y_i \) is the corresponding label. The objective is to minimize the loss function:

\[
\mathcal{L}_{\text{SFT}}(\theta) = \frac{1}{N} \sum_{i=1}^{N} \mathcal{L}(f_\theta(x_i), y_i)
\]

where \( \mathcal{L} \) measures the discrepancy between the model's output and the true label.

\textbf{Gradient Descent Optimization.}
The model's parameters \( \theta \) are updated using gradient-based methods like Stochastic Gradient Descent (SGD) or Adam:

\[
\theta_{t+1} = \theta_t - \eta \nabla_\theta \mathcal{L}_{\text{SFT}}(\theta_t)
\]

SFT is effective for adapting pre-trained models to specific tasks, though it relies on large labeled datasets for fine-tuning.

\subsubsection{SPIN.}
SPIN centers on a self-play mechanism where the LLM simultaneously acts as both the main player and the opponent. During fine-tuning, the main player (the current LLM) is trained to differentiate between data distributions produced by the opponent (the LLM from the prior iteration) and human-annotated target responses, iteratively aligning the LLM with the target data distribution.

\section{Clear View of Model Sequence.}
\label{app:sequence}

We define the models and their training schemes as follows:
\begin{itemize}[topsep=2pt,itemsep=0pt,parsep=0pt]
  \item[\boldmath$M_0$:] Initial supervised fine-tuning (SFT) model.

  \item[\boldmath$M_1$:] Initialized from $M_0$, then fine-tuned on $\mathcal{D}_0$ with:
  \begin{align*}
    &\text{Opponents: } \pi_0(\cdot | \mathbf{x}) \\
    &\text{Main players: } \lambda \cdot \log \frac{\pi_{\boldsymbol\theta}(\cdot | \mathbf{x})}{\pi_{\boldsymbol\theta_0}(\cdot | \mathbf{x})}
  \end{align*}

  \item[\boldmath$M_2$:] Initialized from $M_1$, then fine-tuned on $\mathcal{D}_1, \mathcal{D}_0$ with:
  \begin{align*}
    &\text{Opponents: } \pi_1(\cdot | \mathbf{x}), \pi_0(\cdot | \mathbf{x}) \\
    &\text{Main players: } 
    \begin{aligned}[t]
        &\lambda \cdot \log \frac{\pi_{\boldsymbol\theta}(\cdot | \mathbf{x})}{\pi_{\boldsymbol\theta_1}(\cdot | \mathbf{x})},\\
        &\lambda \cdot \log \frac{\pi_{\boldsymbol\theta}(\cdot | \mathbf{x})}{\pi_{\boldsymbol\theta_0}(\cdot | \mathbf{x})}
    \end{aligned}
  \end{align*}

  \item[\boldmath$M_3$:] Initialized from $M_2$, then fine-tuned on $\mathcal{D}_2, \mathcal{D}_1, \mathcal{D}_0$ with:
  \begin{align*}
    &\text{Opponents: } \pi_2(\cdot | \mathbf{x}), \pi_1(\cdot | \mathbf{x}), \pi_0(\cdot | \mathbf{x}) \\
    &\text{Main players: }
    \begin{aligned}[t]
        &\lambda \cdot \log \frac{\pi_{\boldsymbol\theta}(\cdot | \mathbf{x})}{\pi_{\boldsymbol\theta_2}(\cdot | \mathbf{x})},\\
        &\lambda \cdot \log \frac{\pi_{\boldsymbol\theta}(\cdot | \mathbf{x})}{\pi_{\boldsymbol\theta_1}(\cdot | \mathbf{x})}, \\
        &\lambda \cdot \log \frac{\pi_{\boldsymbol\theta}(\cdot | \mathbf{x})}{\pi_{\boldsymbol\theta_0}(\cdot | \mathbf{x})}
    \end{aligned}
  \end{align*}

  \item[\boldmath$M_4$:] Initialized from $M_3$, then fine-tuned on $\mathcal{D}_3, \mathcal{D}_2, \mathcal{D}_1$ with:
    \begin{align*}
        &\text{Opponents: } \pi_3(\cdot | \mathbf{x}),\ \pi_2(\cdot | \mathbf{x}),\ \pi_1(\cdot | \mathbf{x}) \\
        &\text{Main players: }
        \begin{aligned}[t]
            &\lambda \cdot \log \frac{\pi_{\boldsymbol\theta}(\cdot | \mathbf{x})}{\pi_{\boldsymbol\theta_3}(\cdot | \mathbf{x})}, \\
            &\lambda \cdot \log \frac{\pi_{\boldsymbol\theta}(\cdot | \mathbf{x})}{\pi_{\boldsymbol\theta_2}(\cdot | \mathbf{x})},\\ 
            &\lambda \cdot \log \frac{\pi_{\boldsymbol\theta}(\cdot | \mathbf{x})}{\pi_{\boldsymbol\theta_1}(\cdot | \mathbf{x})}
        \end{aligned}
    \end{align*}

\end{itemize}

\section{Notation Summary}
\label{app:notation}

Table~\ref{tab:notation} summarizes the main symbols used throughout the paper, especially the policy notation across iterations.

\begin{table*}[t]
    \centering
    \small
    \setlength{\tabcolsep}{8pt}
    \renewcommand{\arraystretch}{1.12}
    \begin{tabular}{p{0.2\textwidth} p{0.74\textwidth}}
        \toprule
        \textbf{Notation} & \textbf{Meaning} \\
        \midrule
        $\pi_{\btheta}$ & Current policy model being optimized at the present training step. Only this parameter set is updated by backpropagation. \\
        $\pi_{\btheta_t}$ & Policy checkpoint obtained after completing iteration $t$; also used as the newest opponent model at iteration $t+1$. \\
        $\pi_{\btheta_j}$ & A generic historical checkpoint indexed by $j \in \{t,t-1,t-2\}$, used as a fixed reference model in main-player scoring. \\
        $t$ & Iteration index of team-based self-play. \\
        $\xb, \yb$ & Prompt and target response from the SFT dataset. \\
        $\yb^{\text{gen}}$ & Model-generated response sampled from an opponent policy for prompt $\xb$. \\
        $\mathcal{D}_{\text{SFT}}$ & Original supervised fine-tuning dataset of prompt-response pairs. \\
        $\mathcal{D}_t$ & Triple-form dataset collected at iteration $t$, i.e., $\{(\xb,\yb,\yb^{\text{gen}})\}$. \\
        $\mathcal{D}_O$ & Aggregated opponent dataset formed by recent triple datasets, e.g., $\mathcal{D}_t \cup \mathcal{D}_{t-1} \cup \mathcal{D}_{t-2}$. \\
        $P_j(\xb,\yb)$ & Main-player score defined by the log-ratio between the current policy $\pi_{\btheta}$ and a historical checkpoint $\pi_{\btheta_j}$. \\
        $\mathcal{L}_j$ & Training objective of the $j$-th main player. \\
        $\bm{\alpha}_j$ & Adaptive target-response weight for player $P_j$. \\
        $\bm{\beta}_j$ & Adaptive player weight assigned to the $j$-th main player. \\
        $\lambda$ & Regularization coefficient controlling deviation between the updated policy and the previous checkpoint. \\
        $\bm{\eta}$ & Hyperparameter controlling the strength of target-response reweighting. \\
        $\bm{\gamma}$ & Hyperparameter controlling the sharpness of player-weight assignment. \\
        $N_{\max}$ & Maximum team size, i.e., the number of historical checkpoints retained in the team. \\
        \bottomrule
    \end{tabular}
    \caption{Summary of the main notations used in \ours.}
    \label{tab:notation}
\end{table*}

\begin{table*}[!t]
    \centering
    \footnotesize
    \setlength{\tabcolsep}{3pt}
    \renewcommand{\arraystretch}{0.98}
    \begin{minipage}[t]{0.48\textwidth}
        \centering
        \resizebox{\linewidth}{!}{%
        \begin{tabular}{c|l|c}
            \toprule
            Iteration & Model & Acc~(\%) \\
            \midrule
            - & \texttt{Qwen2.5-1.5B-SFT} & 51.10 \\
            \midrule
            \multirow{6}{*}{Iter 1} & \texttt{SPIN} & 53.75 $_{\scalebox{0.85}{{(+2.65)}}}$ \\
             & \ours & 54.13 $_{\scalebox{0.85}{{(+3.11)}}}$ \\
             & ~~w/o TRW & 54.13 $_{\scalebox{0.85}{{(+3.11)}}}$ \\
             & ~~w/o MPW & 54.13 $_{\scalebox{0.85}{{(+3.11)}}}$ \\
             & ~~w/o TRW+MPW & 54.36 $_{\scalebox{0.85}{{(+3.24)}}}$ \\
             & ~~w/o Team & 53.30 $_{\scalebox{0.85}{{(+2.20)}}}$ \\
            \midrule
            \multirow{6}{*}{Iter 2} & \texttt{SPIN} & 54.36 $_{\scalebox{0.85}{{(+0.61)}}}$ \\
             & \ours & 55.19 $_{\scalebox{0.85}{{(+1.06)}}}$ \\
             & ~~w/o TRW & 56.18 $_{\scalebox{0.85}{{(+2.05)}}}$ \\
             & ~~w/o MPW & 55.65 $_{\scalebox{0.85}{{(+1.52)}}}$ \\
             & ~~w/o TRW+MPW & 55.95 $_{\scalebox{0.85}{{(+1.59)}}}$ \\
             & ~~w/o Team & 55.12 $_{\scalebox{0.85}{{(+1.82)}}}$ \\
            \bottomrule
        \end{tabular}}
    \end{minipage}\hfill
    \begin{minipage}[t]{0.48\textwidth}
        \centering
        \resizebox{\linewidth}{!}{%
        \begin{tabular}{c|l|c}
            \toprule
            Iteration & Model & Acc~(\%) \\
            \midrule
            \multirow{6}{*}{Iter 3} & \texttt{SPIN} & 53.75 $_{\scalebox{0.85}{{(-0.61)}}}$ \\
             & \ours & 56.56 $_{\scalebox{0.85}{{(+1.37)}}}$ \\
             & ~~w/o TRW & 55.42 $_{\scalebox{0.85}{{(-0.76)}}}$ \\
             & ~~w/o MPW & 56.40 $_{\scalebox{0.85}{{(+0.75)}}}$ \\
             & ~~w/o TRW+MPW & 55.42 $_{\scalebox{0.85}{{(-0.53)}}}$ \\
             & ~~w/o Team & 55.19 $_{\scalebox{0.85}{{(-0.62)}}}$ \\
            \midrule
            \multirow{6}{*}{Iter 4} & \texttt{SPIN} & 54.59 $_{\scalebox{0.85}{{(+0.84)}}}$ \\
             & \ours & \textbf{56.94} $_{\scalebox{0.85}{{(+0.38)}}}$ \\
             & ~~w/o TRW & 55.42 $_{\scalebox{0.85}{{(±0.00)}}}$ \\
             & ~~w/o MPW & 55.95 $_{\scalebox{0.85}{{(-0.45)}}}$ \\
             & ~~w/o TRW+MPW & 55.65 $_{\scalebox{0.85}{{(+0.23)}}}$ \\
             & ~~w/o Team & 55.65 $_{\scalebox{0.85}{{(+0.46)}}}$ \\
            \bottomrule
        \end{tabular}}
    \end{minipage}
    \caption{Performance of \ours based on Qwen2.5-1.5B-SFT across GSM8k datasets. Both SPIN and \ours are trained on the GSM8k training set. The numbers in the subscripts represent the accuracy changes compared to the corresponding methods in the previous iteration.}
    \label{tab:ablation}
\end{table*}

\begin{table*}[!t]
    \centering
    \resizebox{\textwidth}{!}{%
    \begin{tabular}{c | c c c c c c | c}
    \toprule
         & \multicolumn{6}{c|}{\textbf{Open LLM Leaderboard V1}} & \\[0.5ex]
        \multirow{-2}{*}{\textbf{Model}} & Arc & TruthfulQA & Winogrande & GSM8k & HellaSwag & MMLU & \multirow{-2}{*}{ \phantom{0} \textbf{Avg.} \phantom{0}} \\
        \midrule
        \texttt{Qwen2.5-1.5B-SFT} & 50.94 & 46.83 & 64.80 & 51.25 & 64.91 & 58.93 & 56.28 \\
        \midrule
        \texttt{Qwen2.5-1.5B-SFT-epoch2} & 51.79 & 45.93 & 62.35 & 49.36 & 65.64 & 57.17 & 55.37 \\
        \texttt{Qwen2.5-1.5B-SFT-epoch3} & 52.47 & 44.87 & 62.12 & 49.43 & 66.70 & 57.31 & 55.48 \\
        \midrule

        \ours Iter-1~(Ours) & 52.99 & 47.64 & 64.96 & 52.77 & 65.77 & 59.50 & 57.27 \\
        \ours Iter-2~(Ours) & 53.41 & 47.91 & 64.17 & 53.15 & 66.08 & 59.58 & 57.38 \\
        \ours Iter-3~(Ours) & 53.67 & 47.91 & 65.04 & 53.60 & 66.18 & 59.52 & 57.65 \\
        \ours Iter-4~(Ours) & 52.56 & 47.90 & 65.11 & 55.04 & 66.30 & 59.64 & \textbf{57.76} \\

    \bottomrule
    \toprule
        & \multicolumn{6}{c|}{\textbf{Open LLM Leaderboard V2}} & \\[0.5ex]
        \multirow{-2}{*}{\textbf{Model}} & IFEval & BBH &  Math & GPQA & MUSR & MMLU-pro & \multirow{-2}{*}{ \phantom{0} \textbf{Avg.} \phantom{0}} \\
        \midrule
        \texttt{Qwen2.5-1.5B-SFT} & 31.73 & 15.15 & 6.65 & 3.47 & 6.02 & 17.38 & 13.40 \\
        \midrule
        \texttt{Qwen2.5-1.5B-SFT-epoch2} & 32.81 & 12.79 & 8.23 & 2.13 & 3.69 & 16.44 & 12.68 \\
        \texttt{Qwen2.5-1.5B-SFT-epoch3} & 33.12 & 14.26 & 7.25 & 4.92 & 2.08 & 16.56 & 13.03 \\
        \midrule
        \ours Iter-1~(Ours) & 32.35 & 15.60 & 8.69 & 3.24 & 5.66 & 17.65 & 13.87 \\
        \ours Iter-2~(Ours) & 35.52 & 15.37 & 8.69 & 4.36 & 5.98 & 17.72 & 14.61 \\
        \ours Iter-3~(Ours) & 36.10 & 15.03 & 10.20 & 4.14 & 5.43 & 18.03 & \textbf{14.82} \\
        \ours Iter-4~(Ours) & 36.04 & 15.51 & 9.37 & 4.36 & 4.73 & 17.82 & 14.64 \\

    \bottomrule
    \end{tabular}%
    }
    \caption{Detailed evaluation data of More Epochs of SFT.\label{tab:furtherdata}}
\end{table*}

\begin{table*}[t]
    \centering
    
    \resizebox{0.9\textwidth}{!}{%
    \begin{tabular}{c | c c c c c c | c}
    \toprule
         & \multicolumn{6}{c|}{\textbf{Open LLM Leaderboard V1}} & \\[0.5ex]
        \multirow{-2}{*}{\textbf{Model}} & Arc & TruthfulQA & Winogrande & GSM8k & HellaSwag & MMLU & \multirow{-2}{*}{ \phantom{0} \textbf{Avg.} \phantom{0}} \\
        \midrule
        \texttt{Qwen2.5-1.5B-SFT} & 50.94±1.46 & 46.83±1.52 & 64.80±1.34 & 51.25±1.38 & 64.91±0.48 & 58.93±0.40 & 56.28 \\
        \midrule
        \ours Iter-4~(Ours) & 52.56±1.46 & 47.90±1.52 & 65.11±1.34 & 55.04±1.37 & 66.30±0.47 & 59.64±0.40 & 57.76 \\
    \bottomrule
    \toprule
        & \multicolumn{6}{c|}{\textbf{Open LLM Leaderboard V2}} & \\[0.5ex]
        \multirow{-2}{*}{\textbf{Model}} & IFEval & BBH &  Math & GPQA & MUSR & MMLU-pro & \multirow{-2}{*}{ \phantom{0} \textbf{Avg.} \phantom{0}} \\
        \midrule
        \texttt{Qwen2.5-1.5B-SFT} & 31.73±1.83 & 15.15±0.60 & 6.65±0.67 & 3.47±1.30 & 6.02±1.70 & 17.38±0.40 & 13.40 \\
        \midrule
        \ours Iter-4~(Ours) & 36.04±1.96 & 15.51±0.60 & 9.37±0.77 & 4.36±1.30 & 4.73±1.66 & 17.82±0.40 & 14.64 \\
    \bottomrule
    \end{tabular}%
    }
     \vspace{-6pt}
    \caption{Example Experiment Results with Standard Error.}
    
    \label{tab:error}
\end{table*}

\section{Additional Experiment Results}

\subsection{Qualitative GSM8K Case Study}
\label{app:gsm8k-case}

To complement the quantitative GSM8K results in the main text, we provide two representative case studies with verbatim model outputs. The first example illustrates a success case where \ours corrects a structural reasoning error made by SPIN. The second example highlights a remaining limitation: \ours improves global reasoning consistency, but does not completely eliminate token-level arithmetic mistakes.

\paragraph{Case 1: Multi-Entity Aggregation Failure.}
We observe a common failure mode in SPIN-style iterative self-training: the model may lose track of earlier entities during the final aggregation step in multi-step reasoning problems.

\noindent\textbf{Problem.}
\begin{quote}
\small
In a neighborhood, the number of rabbit pets is twelve less than the combined number of pet dogs and cats. If there are two cats for every dog, and the number of dogs is 60, how many pets in total are in the neighborhood?
\end{quote}

\noindent\textbf{Ground truth (abridged).}
\begin{quote}
\small
Cats $= 2 \times 60 = 120$ \\
Dogs + Cats $= 180$ \\
Rabbits $= 180 - 12 = 168$ \\
Total $= 168 + 180 = \textbf{348}$
\end{quote}

\noindent\textbf{SPIN output (verbatim).}
\begin{quote}
\small
If there are two cats for every dog, then there are $60 \times 2 = 120$ cats. \\
The combined number of pet dogs and cats is $120 + 60 = 180$ pets. \\
The number of rabbit pets is $180 - 12 = 168$ pets. \\
The total number of pets in the neighborhood is \textbf{$168 + 60 = 228$} pets. \\
\texttt{\#\#\#\# 228}
\end{quote}

\noindent\textbf{TPAW output (verbatim).}
\begin{quote}
\small
If there are two cats for every dog, then there are $60 \times 2 = 120$ cats. \\
The combined number of pet dogs and cats is $120 + 60 = 180$. \\
The number of rabbit pets is $180 - 12 = 168$. \\
The total number of pets in the neighborhood is \textbf{$168 + 60 + 120 = 348$}. \\
\texttt{\#\#\#\# 348}
\end{quote}

\noindent\textbf{Analysis.}
SPIN correctly computes all intermediate quantities, but drops the cat count during the final aggregation step, summing only rabbits and dogs. This reflects a context-dropout phenomenon in iterative self-training, where the model over-relies on the most recent intermediate variable.

\noindent\textbf{Why \ours helps.}
\ours's team-based self-play introduces historical checkpoints as main players. When the current policy begins to neglect earlier entities, gradients from team members that still correctly score the full aggregation provide a corrective signal. This acts as a regularizer against partial forgetting in long reasoning chains, which is difficult to achieve with a self-play trajectory.

\paragraph{Case 2: Token-Level Arithmetic Slip.}
We also note that \ours does not eliminate all GSM8K failure modes.

\noindent\textbf{Problem.}
\begin{quote}
\small
Lloyd has an egg farm. His chickens produce 252 eggs per day and he sells them for \$2 per dozen. How much does Lloyd make on eggs per week?
\end{quote}

\noindent\textbf{Ground truth (abridged).}
\begin{quote}
\small
$252 \times 7 = \textbf{1764}$ eggs per week $\ldots$
\end{quote}

\noindent\textbf{TPAW output (verbatim).}
\begin{quote}
\small
There are 7 days in a week, so Lloyd sells $252 \times 7 = \textbf{1784}$ eggs per week $\ldots$
\end{quote}

\noindent\textbf{Interpretation.}
Here, \ours makes a token-level arithmetic slip ($1764 \rightarrow 1784$) despite following the correct reasoning structure. This suggests that \ours mainly improves global reasoning consistency and entity tracking, but does not directly address low-level arithmetic precision. Addressing such errors likely requires orthogonal techniques such as tool use or arithmetic verification.

\noindent\textbf{Summary.}
These case studies suggest that the GSM8K gains of \ours primarily stem from reducing structural reasoning failures, such as entity omission and aggregation inconsistency, rather than uniformly improving all aspects of numerical accuracy.

\subsection{Further Ablation Study}
\label{app:abalation}

\begin{itemize}[left=0cm]
  \setlength{\itemsep}{3pt}  
  \setlength{\parskip}{2pt}  
  \setlength{\parsep}{2pt}   
  \item Without adaptive Target Response Weighting and adaptive Main Player Weighting (\textbf{w/o TRW+MPW}): both the target response weight and main players weight are fixed.
\end{itemize}

See Table~\ref{tab:ablation} for detailed evaluation data of ablation study.
The experimental results on w/o TRW+MPW are basically consistent with the conclusions of the main text, demonstrate that the dual adaptive weighting mechanism in \ours are critical to achieving optimal performance.

\subsection{Further Experiment Result for More Epochs of SFT}
\label{app:moresft}

As we showed in the main text training for more epochs on the dataset using SFT does not lead to further performance improvements, and may even lead to a decrease in performance. See Table~\ref{tab:furtherdata} for detailed evaluation data.

\subsection{Error Bars of Experiment Result}
\label{app:error}

Here we present a subset of the Standard Error data obtained from our harness-based evaluations in Table~\ref{tab:error}. The relatively low Standard Error values indicate that the evaluation process is stable. Moreover, the Standard Error does not increase noticeably after training, suggesting that the performance improvements are not due to chance but are statistically significant and consistently stable.

\begin{figure}[!htbp]
    \centering
    \includegraphics[width=0.88\linewidth]{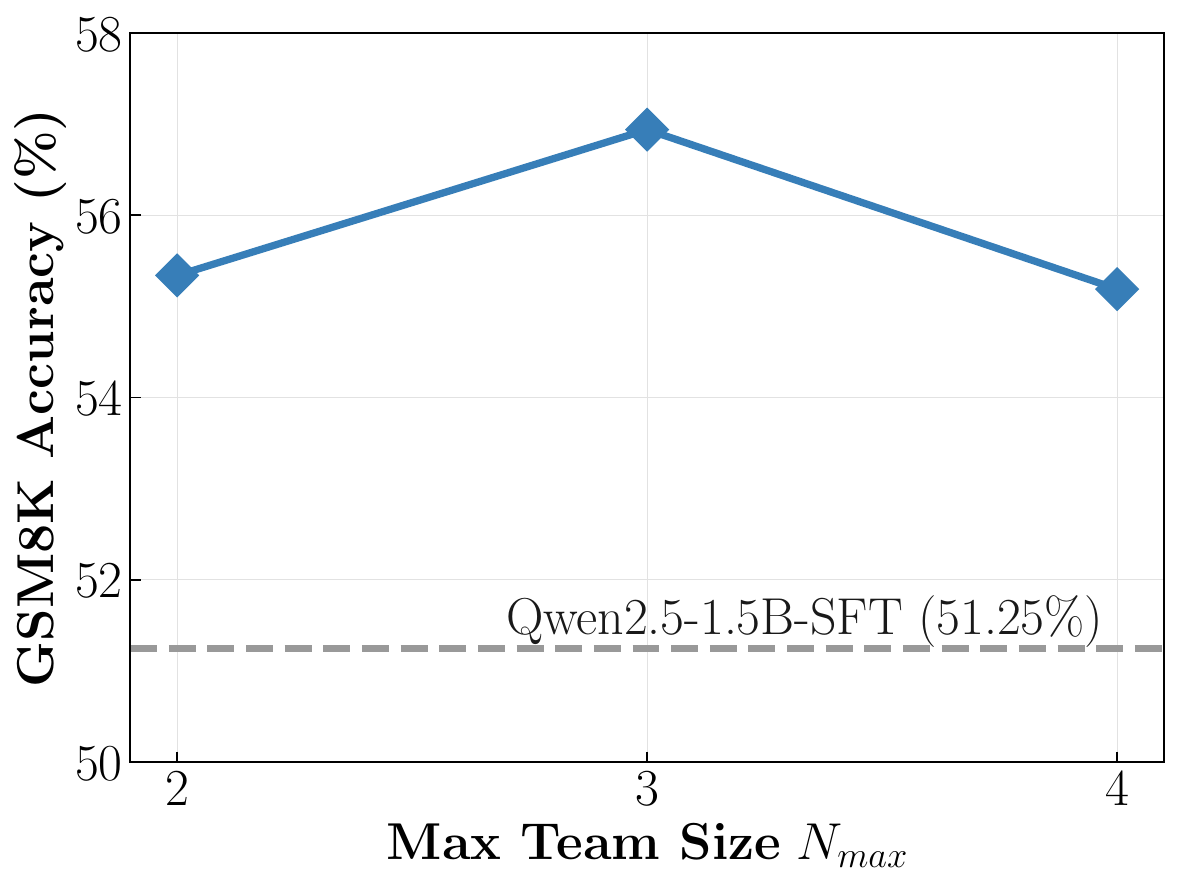}
    \vspace{-5pt}
    \caption{Impact of $N_{\max}$ on GSM8K accuracy.}
    \label{fig:Nmax}
\vspace{0pt}
\end{figure}

\begin{table}[!htbp]
    \centering
    \scriptsize
    \captionsetup{font=footnotesize,skip=2pt}
    \setlength{\tabcolsep}{2pt}
    \renewcommand{\arraystretch}{1.0}
    \begin{subtable}[t]{0.47\linewidth}
        \centering
        \resizebox{\linewidth}{!}{%
        \begin{tabular}{c|ccccc}
            \toprule
            \textbf{$\bm{\gamma}$} & \textbf{0.00} & \textbf{0.25} & \textbf{0.50} & \textbf{0.75} & \textbf{1.00} \\
            \midrule
            Iter-1 & 54.13 & 55.57 & 54.13 & 54.28 & 55.57 \\
            Iter-2 & 55.65 & 55.19 & 55.19 & 55.12 & 55.95 \\
            Iter-3 & 56.40 & 55.19 & 56.56 & 55.34 & 56.03 \\
            Iter-4 & 55.95 & 55.27 & \textbf{56.94} & 55.95 & 55.57 \\
            \bottomrule
        \end{tabular}}
        \caption{Effect of $\bm{\gamma}$.}
        \label{tab:hyper_gamma}
    \end{subtable}\hfill
    \begin{subtable}[t]{0.51\linewidth}
        \centering
        \resizebox{\linewidth}{!}{%
        \begin{tabular}{c|cccccc}
            \toprule
            \textbf{$\bm{\eta}$} & \textbf{1} & \textbf{2} & \textbf{4} & \textbf{6} & \textbf{8} & \textbf{10} \\
            \midrule
            Iter-1 & 54.13 & 54.06 & 54.21 & 54.13 & 54.21 & 54.13 \\
            Iter-2 & 56.18 & 54.82 & 55.95 & 55.19 & 56.53 & 56.33 \\
            Iter-3 & 55.42 & 54.28 & 55.27 & 56.56 & 56.41 & 56.79 \\
            Iter-4 & 55.42 & 54.59 & 55.27 & \textbf{56.94} & 56.71 & 56.63 \\
            \bottomrule
        \end{tabular}}
        \caption{Effect of $\bm{\eta}$.}
        \label{tab:hyper_eta}
    \end{subtable}
    
    \vspace{1pt}
    \begin{subtable}[t]{0.56\linewidth}
        \centering
        \resizebox{\linewidth}{!}{%
        \begin{tabular}{c|ccc}
            \toprule
            \textbf{$N_{max}$} & \textbf{2} & \textbf{3} & \textbf{4} \\
            \midrule
            Iter-3 & 54.89 & 56.56 & 56.56 \\
            Iter-4 & 55.34 & \textbf{56.94} & 55.19 \\
            \bottomrule
        \end{tabular}}
        \caption{Effect of $N_{max}$.}
        \label{tab:hyper_nmax}
    \end{subtable}
    \vspace{-2pt}
    \caption{Detailed GSM8K results for the hyperparameter study. Panels (a)--(c) show the effects of $\bm{\gamma}$, $\bm{\eta}$, and $N_{max}$, respectively.}
    \label{tab:hyperdata}
\end{table}

\vspace{-10pt}

\subsection{Further Experiment Result for Impact of Hyperparameters}
\label{app:hyper}

We further investigate the impact of the team size hyperparameter $N_{\max}$, which denotes the final number of team members and corresponds to the size of the historical checkpoint window considered throughout the game. The experimental results are presented in Figure~\ref{fig:Nmax}.
Notably, when $N_{\max}$ is set to 2 or 4, the observed suboptimal performance may indicate the effects of underfitting and overfitting, respectively.

Additionally, we report comprehensive experimental results for the hyperparameter study in Table~\ref{tab:hyperdata}. These results further demonstrate the soundness of our hyperparameter choices and the robustness of the proposed method.


\end{document}